\documentclass[journal]{IEEEtran}
\usepackage{algorithm}
\usepackage{algorithmic}
\usepackage{multirow}
\usepackage{bbm}
\usepackage{amsmath,amssymb,mathrsfs}
\usepackage{graphicx} %???????
\usepackage{float} %?????????? ?
\usepackage{subfigure} %?????????????
\usepackage{amsthm}
\usepackage[switch]{lineno} 
\usepackage{amsmath,amssymb}
\usepackage{algorithm}
\usepackage{algorithmic}
\usepackage{graphicx}
\usepackage{comment}
\usepackage{color}
\usepackage{subfigure}
\usepackage{multirow}
\usepackage{algorithm}
\usepackage{algorithmic}
\usepackage{wrapfig}
\usepackage{authblk}
\usepackage{booktabs}
\usepackage{hyperref}
\hypersetup{
	colorlinks=true,
	urlcolor=cyan,
}

\ifCLASSINFOpdf
\else
\fi
\begin{document}
	\newtheorem{theorem}{Theorem}
	\newtheorem{lemma}{Lemma}
	\newtheorem{property}{Property}	
	\newtheorem{theorem2}{Theorem}
	\newtheorem{lemma2}{Lemma}
	\newtheorem{property2}{Property}	
	\title{Action Candidate Driven Clipped Double Q-learning for Discrete and Continuous Action Tasks}
	\newcommand*\samethanks[1][\value{footnote}]{\footnotemark[#1]}
	%\author[ ]{Haobo Jiang$^{\dagger}$}
	%\author[ ]{Jianjun Qian$^{\dagger}$}
	\author[ ]{
		Haobo Jiang, Jin Xie, and Jian Yang
		\thanks{Haobo Jiang, Jin Xie and Jian Yang are with PCA Lab, Key Lab of
			Intelligent Perception and Systems for High-Dimensional
			Information of Ministry of Education, and Jiangsu Key Lab of Image
			and Video Understanding for Social Security, School of Computer
			Science and Engineering, Nanjing University of Science and
			Technology, Nanjing, China (E-mail: \{jiang.hao.bo, csjxie, csjyang\}@njust.edu.cn).			
		}}
%	\author[ ]{Haobo Jiang, Jin Xie, Jian Yang}
%	\affil[ ]{PCALab, Nanjing University of Science and Technology, China}
%	\affil[ ]{\tt\small {\{jiang.hao.bo, csjxie, csjyang\}@njust.edu.cn}}
	
	\markboth{Journal of \LaTeX\ Class Files,~Vol.~14, No.~8, August~2015}%
	{Shell \MakeLowercase{\mathrm{et al.}}: Bare Demo of IEEEtran.cls for IEEE Journals}
	\maketitle
	
\begin{abstract}
Double Q-learning is a popular reinforcement learning algorithm in Markov decision process (MDP) problems. 
Clipped Double Q-learning, as an effective variant of Double Q-learning, employs the clipped double estimator to approximate the maximum expected action value.
Due to the underestimation bias of the clipped double estimator, the performance of clipped Double Q-learning may be degraded in some stochastic environments. 
In this paper, in order to reduce the underestimation bias, we propose an action candidate-based clipped double estimator for Double Q-learning.
Specifically, we first select a set of elite action candidates with high action values from one set of estimators.
Then, among these candidates, we choose the highest valued action from the other set of estimators.
Finally, we use the maximum value in the second set of estimators to clip the action value of the chosen action in the first set of estimators and the clipped value is used for approximating the maximum expected action value. 
Theoretically, the underestimation bias in our clipped Double Q-learning decays monotonically as the number of action candidates decreases. 
Moreover, the number of action candidates controls the trade-off between the overestimation and underestimation biases.
In addition, we also extend our clipped Double Q-learning to continuous action tasks via approximating the elite continuous action candidates.
We empirically verify that our algorithm can more accurately estimate the maximum expected action value on some toy environments and yield good performance on several benchmark problems. 
\href{https://github.com/Jiang-HB/AC_CDQ}{Code is available at https://github.com/Jiang-HB/AC\_CDQ}.
	\end{abstract}
	
	\section{Introduction}
In recent years, reinforcement learning has achieved more and more attention.
It aims to learn an optimal policy so that cumulative rewards can be maximized via trial-and-error in an unknown environment \cite{thrun2000reinforcement,kaelbling1996reinforcement,watkins1989learning,arjona2018rudder,szepesvari2010algorithms,wu2021deep}.
Q-learning \cite{watkins1992q,bertsekas1995neuro} is one of the widely studied reinforcement learning algorithms.
As a model-free reinforcement learning algorithm, it generates the optimal policy via selecting the action which owns the largest estimated action value.
In each update, Q-learning executes the maximization operation over action values for constructing the target value of the Q-function.
Unfortunately, this maximization operator tends to overestimate the action values.
Due to the large positive bias, it is difficult to learn the high-quality policy for the Q-learning in many tasks \cite{thrun1993issues,szita2008many,strehl2009reinforcement,van2018deep,van2011insights,xue2021inverse}.
Moreover, such overestimation bias also exists in a variety of variants of Q-learning such as fitted Q-iteration \cite{strehl2006pac,riedmiller2005neural,tosatto2017boosted}, delayed Q-learning \cite{ernst2005tree,abed2018double} and deep Q-network (DQN) \cite{mnih2013playing,mnih2015human,sorokin2015deep,fan2020theoretical}.
	
	Recently, several improved Q-learning methods have been proposed to reduce the overestimation bias.
	Bias-corrected Q-learning \cite{lee2013bias,lee2019bias} adds a bias correction term on the target value  so that the overestimation error can be reduced.
	Softmax Q-learning \cite{song2019revisiting} and Weighted Q-learning \cite{d2016estimating,cini2020deep} are proposed to soften the maximum operation via replacing it with the sum of the weighted action values.
	The softmax operation and Gaussian approximation are employed to generate the weights, respectively. 
	In Averaged Q-learning \cite{anschel2017averaged} and Maxmin Q-learning \cite{lan2020maxmin}, their target values are constructed to reduce the bias and variance via combining multiple Q-functions.
	
	Double Q-learning \cite{hasselt2010double,van2013estimating,zhang2017weighted,chen2021randomized,xiong2020finite} is another popular method to avoid the overestimation bias.
	In Double Q-learning, it exploits the online collected experience sample to randomly update one of two Q-functions.
	In each update, the first Q-function selects the greedy action and the second Q-function evaluates its value.
	Although Double Q-learning can effectively relieve the overestimation bias in Q-learning in terms of the expected value, its target value may occasionally be with the large overestimation bias during the training process. 
	To avoid it, clipped Double Q-learning \cite{fujimoto2018addressing} directly uses the maximum action value of one Q-function to clip the target value of the Double Q-learning.
	Clipping Double Q-learning can be viewed as using the clipped double estimator to approximate the maximum expected value.
	However, the clipped double estimator suffers from the large underestimation bias.
	
	In order to reduce the large negative bias of the clipped double estimator, in this paper, we propose an action candidate based clipped double estimator for Double Q-learning.
	Specifically, we first select a set of action candidates corresponding to high action values in one set of estimators. 
	Then, among these action candidates, we choose the action with the highest value in the other set of estimators. 
	At last, the corresponding action value of the selected action in the first set of estimators clipped by the maximum value in the second set of estimators is used to approximate the maximum expected value.
	Actually, in clipped Double Q-learning, the selected action from one Q-function is independent of the action evaluation in the other Q-function. Thus, the selected action may correspond to the low action value in the second Q-function, which results in the large underestimation.
	By bridging the gap between the action selection and action evaluation from both Q-functions, our action candidate based clipped Double Q-learning can effectively reduce the underestimation bias. 
	Theoretically, the underestimation bias in our clipped Double Q-learning decays monotonically as the number of action candidates decreases. 
	Moreover, the number of action candidates can balance the overestimation bias in Q-learning and the underestimation bias in clipped Double Q-learning.
	Furthermore, we extend our action candidate based clipped Double Q-learning to the deep version. Also, based on the action candidate based clipped double estimator, we propose an effective variant of TD3 \cite{fujimoto2018addressing}, named action candidate based TD3, for the continuous action tasks. 
	Extensive experiments demonstrate that our algorithms can yield good performance on the benchmark problems.
	
The preliminary version of this paper was published in AAAI-2021 and we extend it with in-depth discussions and improvements as below.
Firstly, in addition to the analysis of the expected value of our proposed estimator, we provide the theorem contribution about its variance in Theorem 2 in Section~\ref{sec:3b}. 
Secondly, instead of using a pre-defining number of the action candidates, we further discuss an adaptive selection mechanism about the candidate size for our proposed estimator in Section~\ref{sec:3c} and its RL variant in Section~\ref{sec:4a}. 
Thirdly, we also provide the convergence proof about its simultaneous updating, used in action candidate based TD3, in Theorem 4 in Section~\ref{sec:4b}. 
Finally, we empirically verify the effectiveness of our proposed adaptive selection mechanism and test the bias varying of the learned deep Q-function under different numbers of the action candidates in Section~\ref{sec:5a}.  Also, more plots are added to visualize the experimental results in Section~\ref{sec:5}.

The framework of this paper is orginized as below.  We first introduce the  preliminary knowledge of our method in Section~\ref{sec:2}. Then, we propose our action candidate based clipped double estimator and analyze its theoretical properties  in Section~\ref{sec:3}. Next, we present the RL variants of our estimator for discrete and continuous control tasks in Section~\ref{sec:4}. Finally, we evaluate our method on extensive experiments in Section~\ref{sec:5}.
	
To summarize, our main contributions are as follows:
\begin{itemize}
\item We propose a novel action candidate based clipped double estimator to effectively balance the overestimation bias in the single estimator and the underestimation bias in the clipped double estimator. 

\item We analyze the theoretical properties of our estimator in terms of the bias reduction and the variance balance.

\item We apply our estimator into the clipped Double Q-learning and TD3 for robust discrete and continuous action tasks. Also, we provide the convergence proof about their random updating and simultaneous updating. 

\item Extensive experimental results on the toy tasks and the challenging benchmark problems demonstrate that our estimator and its RL variants can obtain significant performance gain.
\end{itemize}
	
	\section{Background}
	\label{sec:2}
	We model the reinforcement learning problem as an infinite-horizon discounted Markov Decision Process (MDP), which comprises a state space $\mathcal{S}$, a discrete action space $\mathcal{A}$, a state transition probability distribution $\mathcal{P}:\mathcal{S} \times \mathcal{A} \times \mathcal{S} \rightarrow \mathbb{R}$, a reward function $R: \mathcal{S}\times\mathcal{A}\rightarrow\mathbb{R}$ and a discount factor $\gamma \in [0,1]$.
	At each step $t$, with a given state $\mathbf{s}_t \in \mathcal{S}$, the agent receives a reward ${r}_t$ and the new state $\mathbf{s}_{t+1} \in \mathcal{S}$ after taking an action $\mathbf{a}_t \in \mathcal{A}$.
	The goal of the agent is to find a policy $\pi: \mathcal{S}\times\mathcal{A}\rightarrow [0, 1]$ that maximizes the accumulated rewards.
	
	In the standard MDP problem, the action value function ${Q^\pi\left(\mathbf{s},\mathbf{a}\right)=\mathbb{E}_\pi\left[\sum_{t=0}^{\infty}\gamma^t r_t\mid\mathbf{s}_0=\mathbf{s}, \mathbf{a}_0=\mathbf{a}\right]}$ denotes the expected return after doing the action $\mathbf{a}$ in the state $\mathbf{s}$ with the policy $\pi$. 
	The optimal policy can be defined as: $\pi^*(\mathbf{s}) = \arg\max_{\mathbf{a}\in \mathcal{A}}Q^*(\mathbf{s},\mathbf{a})$ where the optimal action value function $Q^*\left(\mathbf{s},\mathbf{a}\right)$ satisfies the \textit{Bellman optimality equation}:
	\begin{equation} \label{policy}
		\begin{split}
			Q^*\left(\mathbf{s},\mathbf{a}\right)=\mathbb{E}_{\mathbf{s}'\sim\mathcal{P}\left(\cdot\mid\mathbf{s},\mathbf{a}\right)}\left[R\left(\mathbf{s},\mathbf{a}\right) +\max_{\mathbf{a}'\in\mathcal{A}}Q^*\left(\mathbf{s}',\mathbf{a}'\right)\right]. %\sum_{\mathbf{s}'\in\mathcal{S}}\mathcal{P}\left({\mathbf{s}'}\mid{\mathbf{s},\mathbf{a}}\right)\max_{\mathbf{a}'\in\mathcal{A}}Q^*\left(\mathbf{s}',\mathbf{a}'\right).
		\end{split}
	\end{equation}
	% \begin{equation} \label{policy}
	% \begin{split}
	% Q^*\left(\mathbf{s},\mathbf{a}\right)=R\left(\mathbf{s},\mathbf{a}\right) + \gamma \sum_{\mathbf{s}'\in\mathcal{S}}\mathcal{P}\left({\mathbf{s}'}\mid{\mathbf{s},\mathbf{a}}\right)\max_{\mathbf{a}'\in\mathcal{A}}Q^*\left(\mathbf{s}',\mathbf{a}'\right).
	% \end{split}
	% \end{equation}
	
	\subsection{Clipped Double Q-learning}
	%\textbf{(Double) Q-learning.} 
	To approximate the optimal action value function $Q^*\left(\mathbf{s},\mathbf{a}\right)$, Q-learning constructs a Q-function $Q: \mathcal{S}\times\mathcal{A}\rightarrow\mathbb{R}$ and iteratively updates it in each step via the updating formula:
	\begin{equation}
			 Q\left(\mathbf{s}_t,\mathbf{a}_t\right)\leftarrow Q\left(\mathbf{s}_t,\mathbf{a}_t\right)+ \alpha\left(y_t^\mathrm{Q} - Q\left(\mathbf{s}_t,\mathbf{a}_t\right)\right),
	\end{equation}
	where the target value $y_t^{\mathrm{Q}}$ is defined as below:
	\begin{equation} \label{q} 
		\begin{split}
			y_t^{\mathrm{Q}} = r_{t}+\gamma \max_{\mathbf{a}' \in \mathcal{A}}Q\left(\mathbf{s}_{t+1},\mathbf{a}'\right).
		\end{split}
	\end{equation}
	To relieve the overestimation bias in the Q-learning, Double Q-learning maintains two Q-functions, $Q^\mathrm{A}$ and $Q^\mathrm{B}$, and randomly updates one Q-function, such as $Q^\mathrm{A}$, with the target value $y_t^{\mathrm{DQ}}$ as below:
	\begin{equation} \label{policy}
		\begin{split}
			y_t^{\mathrm{DQ}} = r_t + \gamma Q^\mathrm{B}\left(\mathbf{s}_{t+1}, \arg \max_{\mathbf{a}'\in \mathcal{A}}  Q^\mathrm{A}(\mathbf{s}_{t+1},\mathbf{a}')\right).
		\end{split}
	\end{equation}
	Finally, clipped Double Q-learning uses the maximum action value of one Q-function to clip the target value in Double Q-learning as below to update the Q-function:
	\begin{equation} \label{policy}
		\begin{split}
			y_t^{\mathrm{CDQ}} = r_t + \gamma  \min\left\{Q^\mathrm{A}(\mathbf{s}_{t+1},\mathbf{a}^*), Q^\mathrm{B}(\mathbf{s}_{t+1},\mathbf{a}^*\right\},
		\end{split}
	\end{equation}where $\mathbf{a}^*=\arg\max_{\mathbf{a}}Q^\mathrm{A}\left(\mathbf{s}_{t+1},\mathbf{a}\right)$. As demonstrated in \cite{fujimoto2018addressing}, clipped Double Q-learning can further reduce the chance of the overestimation risk in original Double Q-learning.
	% \subsection{Clipped Double Q-learning.}
	% It uses the maximum action value of one Q-function to clip the target value in Double Q-learning as below to update the Q-function:
	% \begin{equation} \label{policy}
	% \begin{split}
	% y_t^{CDQ} = r_t + \gamma  \min\left\{Q^\mathrm{A}(\mathbf{s}_{t+1},\mathbf{a}^*), Q^\mathrm{B}(\mathbf{s}_{t+1},\mathbf{a}^*\right\},
	% \end{split}
	% \end{equation}where $\mathbf{a}^*=\arg\max_aQ^\mathrm{A}\left(\mathbf{s}_{t+1},\mathbf{a}\right)$.
	
	\subsection{Twin Delayed Deep Deterministic Policy Gradient} 
	Twin Delayed Deep Deterministic policy gradient (TD3) applies the clipped Double Q-learning into the continuous action control under the actor-critic framework.
	Specifically, it maintains an actor network $\mu\left(\mathbf{s}; {\boldsymbol{\phi}}\right)$ and two critic networks $Q\left(\mathbf{s}, \mathbf{a};\boldsymbol{\theta}_1\right)$ and $Q\left(\mathbf{s}, \mathbf{a};\boldsymbol{\theta}_2\right)$. Two critic networks are updated via $\boldsymbol{\theta}_i \leftarrow \boldsymbol{\theta}_i + \alpha \nabla_{\boldsymbol{\theta}_i}\mathbb{E}\left[\left(Q\left(\mathbf{s}_t, \mathbf{a}_t;\boldsymbol{\theta}_i\right)-y_t^{\mathrm{TD3}}\right)^2\right]$. The target value $y_t^{\mathrm{TD3}}$ is defined as below:
	\begin{equation} \label{policy}
		\begin{split}
			y_t^{\mathrm{TD3}} = r_t + \gamma \min_{i=1,2}Q\left(\mathbf{s}_{t+1}, \mu\left(\mathbf{s}_{t+1}; {\boldsymbol{\phi}^-}\right);\boldsymbol{\theta}_i^-\right),
		\end{split}
	\end{equation}
	where $\boldsymbol{\phi}^-$ and $\boldsymbol{\theta}_i^-$ are the soft updated parameters of $\boldsymbol{\phi}$ and $\boldsymbol{\theta}_i$. The actor $\mu\left(\mathbf{s}; {\boldsymbol{\phi}}\right)$ is updated via $\boldsymbol{\phi} \leftarrow \boldsymbol{\phi} + \alpha \nabla_{\boldsymbol{\phi}}J$, where the policy gradient $\nabla_{\boldsymbol{\phi}}J$ is:
	\begin{equation} \label{policy}
		\begin{split}
			\nabla_{\boldsymbol{\phi}}J=\mathbb{E}\left[\left.\nabla_{\mathbf{a}} Q\left(\mathbf{s}_{t}, \mathbf{a} ; \boldsymbol{\theta}_1\right)\right|_{\mathbf{a}=\mu\left(\mathbf{s}_{t} ; \boldsymbol{\phi}\right)} \nabla_{\boldsymbol{\phi}} \mu\left(\mathbf{s}_{t} ; \boldsymbol{\phi}\right)\right].
		\end{split}
	\end{equation}
	
	% \subsection{Distributional Reinforcement Learning} 
	% Distributional RL is devoted to inferring the optimal decision by estimating the value distribution of the random return $Z^\pi\left(\mathbf{s},\mathbf{a}\right)=\sum_{t=0}^{\infty}\gamma^t R\left(\mathbf{s}_t,\mathbf{a}_t\right)$ (where $\mathbf{s}_0=\mathbf{s},\mathbf{a}_0=\mathbf{a}$) rather than the action value $Q^\pi\left(\mathbf{s},\mathbf{a}\right)={\mathbb{E}_\pi\left[Z^\pi\left(\mathbf{s},\mathbf{a}\right)\right]}$ in traditional RL. 
	% In QR-DQN, the random return is approximated by a parameterized quantile distribution $Z_\theta\left(\mathbf{s},\mathbf{a}\right)=\frac{1}{M}\sum_{m=1}^M\delta\left(\theta_\psi^m\left(\mathbf{s},\mathbf{a}\right)\right)$ that is a uniform distribution supported on the locations $\left\{\theta_\psi^1\left(\mathbf{s},\mathbf{a}\right), \ldots, \theta_\psi^M\left(\mathbf{s},\mathbf{a}\right)\right\}$ predicted by the model $\theta_\psi:\mathcal{S}\times\mathcal{A}\rightarrow\mathbb{R}^M$.

	% Specifically, the value distribution of the policy $\pi$ can be approximated via the dynamic planning with the distributional Bellman operator $\mathcal{T}^{\pi}$:
	% \begin{equation}
	%     \mathcal{T}^{\pi}Z\left(\mathbf{s},\mathbf{a}\right)=R\left(\mathbf{s},\mathbf{a}\right)+\gamma Z\left(\mathbf{s}',\mathbf{a}'\right),
	% \end{equation}where $\mathbf{s}'\sim\mathcal{P}\left(\cdot\mid{\mathbf{s},\mathbf{a}}\right)$ and $\mathbf{a}'\sim\pi\left(\cdot\mid\mathbf{s}'\right)$. QR-DQN approximates the 
	
	\begin{algorithm*}[ht]
		\caption{Action Candidate Based Clipped Double Q-learning}
		\label{alg:algorithm1}
		\textbf{Initialize} Q-functions $Q^\mathrm{A}$ and $Q^\mathrm{B}$, initial state $\mathbf{s}$ and buffer $\mathcal{O}$.
		\begin{algorithmic}[0] %[1] enables line numbers
			\REPEAT
			\STATE Select action $\mathbf{a}$ based on $Q^\mathrm{A}\left(\mathbf{s},\cdot\right)$, $Q^\mathrm{B}\left(\mathbf{s},\cdot\right)$ (e.g., $\epsilon$-greedy in $Q^\mathrm{A}\left(\mathbf{s},\cdot\right)+Q^\mathrm{B}\left(\mathbf{s},\cdot\right)$) and observe reward $r$, next state $\mathbf{s}'$.
			%\STATE Update either $Q^\mathrm{A}$ or $Q^\mathrm{B}$ randomly.
			\IF{update $Q^\mathrm{A}$}
			\STATE Add $\hat{\mathcal{H}}\left(Q^\mathrm{A},\mathbf{s}'\right)= \left|Q^\mathrm{A}\left(\mathbf{s}',\mathbf{a}^*\right)-Q^\mathrm{A}\left(\mathbf{s}',\mathbf{a}_L\right)\right|$ into buffer $\mathcal{O}$.
			\STATE Calculate $K=\hat{K}'\left(Q^\mathrm{A},\mathbf{s}',\bar{c}\right)$ where $\bar{c}$ is the mean of the latest $M$ values in buffer $\mathcal{O}$.
			\STATE Determine action candidates ${\mathcal{M}}_K$ from $Q^\mathrm{B}\left(\mathbf{s}', \cdot\right)$ and define $\mathbf{a}_K^* = \arg \max_{\mathbf{a} \in {\mathcal{M}}_K} Q^\mathrm{A}\left(\mathbf{s}', \mathbf{a}\right)$.
			\STATE $Q^\mathrm{A}\left(\mathbf{s},\mathbf{a}\right) \leftarrow Q^\mathrm{A}\left(\mathbf{s},\mathbf{a}\right) + \alpha\left(\mathbf{s},\mathbf{a}\right) \cdot \left( r+\gamma \min\left\{Q^\mathrm{B}\left(\mathbf{s}', \mathbf{a}_K^*\right), {\max}_{\mathbf{a}} Q^\mathrm{A}\left(\mathbf{s}', \mathbf{a}\right)\right\} -Q^\mathrm{A}\left(\mathbf{s},\mathbf{a}\right)\right)$.
			\ELSIF{update $Q^\mathrm{B}$}
			\STATE Add $\hat{\mathcal{H}}\left(Q^\mathrm{B},\mathbf{s}'\right)= \left|Q^\mathrm{B}\left(\mathbf{s}',\mathbf{a}^*\right)-Q^\mathrm{B}\left(\mathbf{s}',\mathbf{a}_L\right)\right|$ into buffer $\mathcal{O}$.
			\STATE Calculate $K=\hat{K}'\left(Q^\mathrm{B},\mathbf{s}',\bar{c}\right)$ where $\bar{c}$ is the mean of the latest $M$ values in buffer $\mathcal{O}$.
			\STATE Determine action candidates ${\mathcal{M}}_K$ from $Q^\mathrm{A}\left(\mathbf{s}', \cdot\right)$ and define $\mathbf{a}_K^* = \arg \max_{\mathbf{a} \in {\mathcal{M}}_K} Q^\mathrm{B}\left(\mathbf{s}', \mathbf{a}\right)$.
			\STATE $Q^\mathrm{B}\left(\mathbf{s},\mathbf{a}\right) \leftarrow Q^\mathrm{B}\left(\mathbf{s},\mathbf{a}\right) + \alpha\left(\mathbf{s},\mathbf{a}\right) \cdot \left(r+\gamma \min\left\{Q^\mathrm{A}\left(\mathbf{s}', \mathbf{a}_K^*\right), {\max}_{\mathbf{a}} Q^\mathrm{B}\left(\mathbf{s}', \mathbf{a}\right)\right\} -Q^\mathrm{B}\left(\mathbf{s},\mathbf{a}\right)  \right)$.
			\ENDIF
			\STATE $\mathbf{s} \leftarrow \mathbf{s}'$
			\UNTIL{end}
		\end{algorithmic}
	\end{algorithm*}

	\section{Estimating the Maximum Expected Value}
	\label{sec:3}
	\subsection{Revisiting the Clipped Double Estimator}
	Suppose that there is a finite set of $N$ $\left(N\geq2\right)$ independent random variables $\mathcal{X}=\left\{X_1,\ldots,X_N\right\}$ with the expected values ${\mu}=\left\{\mu_1,\mu_2,\ldots,\mu_N\right\}$. 
	We consider the problem of approximating the maximum expected value of the variables in $\mathcal{X}$: $\mu^*={\max}_{i}\mu_i={\max}_{i}\mathbb{E}\left[X_i\right]$.
	The clipped double estimator \cite{fujimoto2018addressing} denoted as $\hat{\mu}_{\mathrm{CDE}}^*$ is an effective estimator to estimate the maximum expected value.
	
	Specifically, let $S=\bigcup_{i=1}^{N}S_i$ denotes a set of samples, where $S_i$ is the subset containing samples for the variable $X_i$. 
	We assume that the samples in $S_i$ are independent and identically distributed (i.i.d.). 
	Then, we can obtain a set of the unbiased estimators $\hat{{\mu}}=\left\{\hat{\mu}_1,\hat{\mu}_2,\ldots,\hat{\mu}_N\right\}$ where each element $\hat{\mu}_i$ is an unbiased estimator of $\mathbb{E}\left[X_i\right]$ and can be obtained by calculating the sample average: $\mathbb{E}\left[X_i\right]\approx\hat{\mu}_i \stackrel{\text { def }}{=} \frac{1}{|S_i|}\sum_{s \in S_i}s$. 
	Furthermore, we randomly divide the set of samples $S$ into two subsets: $S^\mathrm{A}$ and $S^\mathrm{B}$. Analogously, two sets of unbiased estimators $\hat{{\mu}}^\mathrm{A}=\left\{\hat{\mu}_1^\mathrm{A},\hat{\mu}_2^\mathrm{A},\ldots,\hat{\mu}_N^\mathrm{A}\right\}$ and $\hat{{\mu}}^\mathrm{B}=\left\{\hat{\mu}_1^\mathrm{B},\hat{\mu}_2^\mathrm{B},\ldots,\hat{\mu}_N^\mathrm{B}\right\}$ can be obtained by sample average:
	$\hat{\mu}_i^\mathrm{A}=\frac{1}{|S_i^\mathrm{A}|}\sum_{s \in S_i^\mathrm{A}}s$, $\hat{\mu}_i^\mathrm{B}=\frac{1}{|S_i^\mathrm{B}|}\sum_{s \in S_i^\mathrm{B}}s$.
	Finally, the clipped double estimator combines the estimator sets $\hat{{\mu}}$, $\hat{{\mu}}^\mathrm{A}$ and $\hat{{\mu}}^\mathrm{B}$ to construct the following estimator, denoted as $\hat{\mu}_{\mathrm{CDE}}^*$, to approximate the maximum expected value:
	\begin{normalsize}
		\begin{equation} \label{cde}
			\begin{split}
				\mu^*={\max}_{i}\mu_i\approx \min\left\{\hat{\mu}_{a^*}^\mathrm{B}, \max_i\hat{\mu}_i\right\},
			\end{split}
		\end{equation}
	\end{normalsize}where the variable $\max_i\hat{\mu}_i$ is called the single estimator denoted as $\hat{\mu}_{\mathrm{SE}}^*$ and the variable $\hat{\mu}_{a^*}^\mathrm{B}$ is called the double estimator denoted as $\hat{\mu}_{\mathrm{DE}}^*$. 

For the single estimator, it directly uses the maximum value of $\hat{{\mu}}$ to approximate the maximum expected value.
Since the expected value of the single estimator is no less than $\mu^*$, the single estimator has an overestimation bias.
Instead, for double estimator, it first calculates the index $a^*$ corresponding to the maximum value in $\hat{{\mu}}^\mathrm{A}$, that is $\hat{{\mu}}_{a^*}^\mathrm{A}=\max_i \hat{{\mu}}_i^\mathrm{A}$, and then uses the value $\hat{\mu}_{a^*}^\mathrm{B}$ to estimate the maximum expected value. 
Due to the expected value of the double estimator is no more than $\mu^*$, it is underestimated.

Although the double estimator is underestimated in terms of the expected value, it still can't completely eliminate the overestimation \cite{fujimoto2018addressing}. By clipping the double estimator via a single estimator, the clipped double estimator can effectively relieve it.
However, due to the expected value of $\min\left\{\hat{\mu}_{a^*}^\mathrm{B}, \max_i\hat{\mu}_i\right\}$ is no more than that of $\hat{\mu}_{a^*}^\mathrm{B}$, the clipped double estimator may further exacerbate the underestimation bias in the double estimator and thus suffer from larger underestimation bias. %Thus, the clipped double estimator suffers from the large underestimation bias.
	
	\begin{algorithm*}[ht]
		\caption{Action Candidate Based TD3}
		\label{actd3}
		Initialize critic networks $Q\left(\cdot;\boldsymbol{\theta}_1\right)$, $Q\left(\cdot;\boldsymbol{\theta}_2\right)$, and actor networks $\mu\left(\cdot; {\boldsymbol{\phi}_1}\right)$,  $\mu\left(\cdot; {\boldsymbol{\phi}_2}\right)$ with random parameters $\boldsymbol{\theta}_1$, $\boldsymbol{\theta}_2$, $\boldsymbol{\phi}_1$, $\boldsymbol{\phi}_2$  \\
		Initialize target networks $\boldsymbol{\theta}_1^- \leftarrow \boldsymbol{\theta}_1$, $\boldsymbol{\theta}_2^- \leftarrow \boldsymbol{\theta}_2$, $\boldsymbol{\phi}_1^- \leftarrow \boldsymbol{\phi}_1$, $\boldsymbol{\phi}_2^- \leftarrow \boldsymbol{\phi}_2$\\
		Initialize replay buffer $\mathcal{D}$
		\begin{algorithmic}[0] %[1] enables line numbers
			\FOR{$t = 1:T$}
			\STATE Select action with exploration noise $\mathbf{a} \sim \mu\left(\mathbf{s}; {\boldsymbol{\phi}_1}\right)+\boldsymbol{\epsilon}$, $\boldsymbol{\epsilon} \sim \mathcal{N}(0, \boldsymbol{\sigma})$  and observe reward  $r$ and next state $\mathbf{s}'$.
			\STATE Store transition tuple $\left\langle \mathbf{s}, \mathbf{a}, r, \mathbf{s}'\right\rangle$ in $\mathcal{D}$.
			\STATE Sample a mini-batch of transitions $\left\{\left\langle \mathbf{s}, \mathbf{a}, r, \mathbf{s}'\right\rangle\right\}$ from $\mathcal{D}$.
			\STATE Determine ${\mathcal{M}}_K = \left\{\mathbf{a}_i \right\}_{i=1}^K, \mathbf{a}_i \sim \mathcal{N}\left(\mu\left(\mathbf{s}'; {\boldsymbol{\phi}_2^-}\right), \bar{\boldsymbol{\sigma}}\right)$ and define $\mathbf{a}_K^* = \arg \max_{\mathbf{a} \in {\mathcal{M}}_K} Q\left(\mathbf{s}', \mathbf{a};\boldsymbol{\theta}_1^-\right)$.
			\STATE Update $\boldsymbol{\theta}_{i} \leftarrow \operatorname{argmin}_{\boldsymbol{\theta}_{i}} N^{-1} \sum{\left[r + \gamma \min\left\{Q\left(\mathbf{s}', \mathbf{a}_K^*;\boldsymbol{\theta}_2^-\right), Q\left(\mathbf{s}', \mu\left(\mathbf{s}'; {\boldsymbol{\phi}_1^-}\right);\boldsymbol{\theta}_1^-\right)\right\}-Q\left(\mathbf{s}, \mathbf{a}; \boldsymbol{\theta}_i\right)\right]}^{2}$.
			\IF{$t$ mod $d$}
			\STATE Update $\boldsymbol{\phi}_i$ by the deterministic policy gradient: $\nabla_{\boldsymbol{\phi}_i} J(\boldsymbol{\phi}_i)=\left.\frac{1}{N} \sum \nabla_{\mathbf{a}} Q_{\theta_{i}}\left(\mathbf{s}, \mathbf{a}\right)\right|_{\mathbf{a}=\mu\left(\mathbf{s}; {\boldsymbol{\phi}_i}\right)} \nabla_{\boldsymbol{\phi}_i} \mu\left(\mathbf{s}; {\boldsymbol{\phi}_i}\right)$.
			\STATE Update target networks: $\boldsymbol{\theta}_{i}^{-} \leftarrow \tau \boldsymbol{\theta}_{i}+(1-\tau) \boldsymbol{\theta}_{i}^{-}$, $\boldsymbol{\phi}_i^{-} \leftarrow \tau \boldsymbol{\phi}_i+(1-\tau) \boldsymbol{\phi}_i^{-}$.
			\ENDIF
			\ENDFOR
		\end{algorithmic}
	\end{algorithm*}
	
	\subsection{Action Candidate Based Clipped Double Estimator}
	\label{sec:3b}
	The double estimator is essentially an underestimated estimator, leading to the underestimation bias. The clipping operation in the clipped double estimator further exacerbates the underestimation problem. Therefore, although the clipped double estimator can effectively avoid the positive bias, it generates a large negative bias. 
	
	In order to reduce the negative bias of the clipped double estimator, we propose an action candidate based clipped double estimator denoted as $\hat{\mu}_{\mathrm{AC}}^*$.
	Notably, the double estimator chooses the index $a^*$  only from the estimator set $\hat{{\mu}}^\mathrm{A}$ and ignores the other estimator set $\hat{{\mu}}^\mathrm{B}$. Thus, it may choose the index $a^*$ associated with the low value in $\hat{{\mu}}^\mathrm{B}$ and generate the small estimation $\hat{\mu}_{a^*}^\mathrm{B}$, leading to the large negative bias.
	Different from the double estimator, instead of selecting the index $a^*$ from $\hat{{\mu}}^\mathrm{A}$ among all indexes, we just choose it from an index subset called candidates. 
	The set of candidates, denoted as ${\mathcal{M}}_K$, is defined as the index subset corresponding to the largest $K$ values in $\hat{{\mu}}^\mathrm{B}$, that is:
		\begin{equation} \label{policy}
			\begin{split}
				{\mathcal{M}}_K = \left\{i\mid\hat{\mu}_i^\mathrm{B} \in {\rm\ top} \ K \ {\rm values \ in \ \hat{{\mu}}^\mathrm{B}} \right\}.
			\end{split}
		\end{equation}
	The variable $a_K^*$ is then selected as the index to maximize $\hat{{\mu}}^\mathrm{A}$ among the index subset $\mathbf{\mathcal{M}}_K$: $\hat{\mu}_{a_K^*}^\mathrm{A}=\max_{i \in \mathbf{\mathcal{M}}_K}\hat{\mu}_i^\mathrm{A}$. 
	If there are multiple indexes owning the maximum value, we randomly pick one. 
	Finally, by clipping, we estimate the maximum expected value as below:
	\begin{normalsize}
		\begin{equation} \label{estimator}
			\begin{split}
				\mu^*={\rm max}_{i}\mu_i={\rm max}_{i}\mathbb{E}\left[\hat{\mu}_i^\mathrm{B}\right] \approx \min\left\{ \hat{\mu}_{a_K^*}^\mathrm{B}, \hat{\mu}_{\mathrm{SE}}^*\right\}.
			\end{split}
		\end{equation}
	\end{normalsize}
	\begin{property}
		%Assume that the values in $\hat{\boldsymbol{\mu}}^\mathrm{A}$ and $\hat{\boldsymbol{\mu}}^\mathrm{B}$ are different. 
		Let $a_K^*$ be the index that maximizes $\hat{{\mu}}^\mathrm{A}$ among $\mathcal{{M}}_K$: $\hat{\mu}_{a_K^*}^\mathrm{A}=\max_{i \in \mathcal{{M}}_K}\hat{\mu}_i^\mathrm{A}$.
		Then, as the number $K$ decreases, the underestimation bias decays monotonically, that is $\mathbb{E}\left[\hat{\mu}_{a_{K}^*}^\mathrm{B}\right] \geq \mathbb{E}\left[\hat{\mu}_{a_{K+1}^*}^\mathrm{B}\right]$, $1\leq K < N$. Moreover, $\forall{K}: 1\leq K \leq N$, $\mathbb{E}\left[\hat{\mu}_{a_K^*}^\mathrm{B}\right] \geq \mathbb{E}\left[\hat{\mu}_{\mathrm{DE}}^*\right]$.
		 %Furthermore, the inequality is strict if and only if $P\left(\hat{\mu}_{a_{(K+1)}}^\mathrm{A} < \hat{\mu}_{a_K^*}^\mathrm{A}\right)=0$
	\end{property}

	For the proof please refer to Appendix I. Consequently, based on Property 1, we further theoretically analyze the estimation bias of action candidate based clipped double estimator as below (proof can be seen in Appendix I).
	\begin{theorem} 
		%	Assume that the values in $\hat{\boldsymbol{\mu}}^\mathrm{A}$ and $\hat{\boldsymbol{\mu}}^\mathrm{B}$ are different.
		As the number $K$ decreases, the underestimation decays monotonically, that is $\mathbb{E}\left[\min\left\{\hat{\mu}_{a_{K}^*}^\mathrm{B}, \hat{\mu}^*_{\mathrm{SE}}\right\}\right] \geq \mathbb{E}\left[\min\left\{\hat{\mu}_{a_{K+1}^*}^\mathrm{B}, \hat{\mu}^*_{\mathrm{SE}}\right\}\right]$, $1\leq K < N$, where the inequality is strict if and only if $P\left( \hat{\mu}_{\mathrm{SE}}^* > \hat{\mu}_{a_{K}^*}^\mathrm{B}> \hat{\mu}_{a_{K+1}^*}^\mathrm{B}\right)>0$ or $P\left(\hat{\mu}_{a_{K}^*}^\mathrm{B} \geq \hat{\mu}_{\mathrm{SE}}^*>\hat{\mu}_{a_{K+1}^*}^\mathrm{B}\right)>0$. Moreover, $\forall{K}: 1\leq K \leq N$, $\mathbb{E}\left[\min\left\{\hat{\mu}_{a_K^*}^\mathrm{B}, \hat{\mu}_{\mathrm{SE}}^*\right\}\right] \geq \mathbb{E}\left[\hat{\mu}_{\mathrm{CDE}}^*\right]$.
	\end{theorem}

	Notably, from the last inequality in Theorem 1, one can see that our estimator can effectively reduce the large underestimation bias in clipped double estimator. Moreover, since the existed inequality $\mathbb{E}\left[\hat{\mu}_{\mathrm{SE}}^*\right]\geq\mathbb{E}\left[\min\left\{ \hat{\mu}_{a_K^*}^\mathrm{B},\hat{\mu}_{\mathrm{SE}}^*\right\}\right]\geq\mathbb{E}\left[\hat{\mu}_{\mathrm{CDE}}^*\right]$,
	it essentially implies that the choice of $K$ controls the trade-off between the overestimation bias in single estimator and the underestimation bias in clipped double estimator. %The proof can be seen in appendix.
	
	%Notably, the clipping operation is necessary for our improved clipped double estimator.
	The upper bound of 
	$\mathbb{E}\left[\hat{\mu}_{\mathrm{SE}}^{*}\right]$ \cite{van2013estimating} is:
	% \begin{small}
	\begin{equation} \label{policy}
		\begin{split}
			\mathbb{E}\left[\hat{\mu}_{\mathrm{SE}}^{*}\right] = \mathbb{E}\left[ {\rm max}_i\hat{\mu}_i \right]  \leq \mu^* + \sqrt{\frac{N-1}{N} \sum_{i}^{N} \operatorname{Var}\left[\hat{\mu}_{i}\right]}.
		\end{split}
	\end{equation}
	% \end{small}
	Since $\mathbb{E}\left[\hat{\mu}_{a_{K}^*}^\mathrm{B}\right]$ decreases monotonically as the number $K$ increases (see Property 1), $\mathbb{E}\left[\hat{\mu}_{a_{1}^*}^\mathrm{B}\right]$ is maximum.
	Due to the candidate subset $\mathcal{M}_1$ only contains one candidate corresponding to the largest value in $\hat{{\mu}}^\mathrm{B}$, we can obtain $\mathbb{E}\left[\hat{\mu}_{a_{1}^*}^\mathrm{B}\right] = \mathbb{E}\left[\max_i\hat{\mu}_i^\mathrm{B}\right]$.
	Similar to the upper bound in $\mathbb{E}\left[\hat{\mu}_{\mathrm{SE}}^{*}\right]$, we can see that $\mathbb{E}\left[\max_i\hat{\mu}_i^\mathrm{B}\right] \leq \mu^* + \sqrt{\frac{N-1}{N} \sum_{i}^{N} \operatorname{Var}\left[\hat{\mu}_{i}^\mathrm{B}\right]}$.
	Since $\hat{\mu}_{i}^\mathrm{B}$ is just estimated via $S_i^\mathrm{B}$ containing half of samples rather than $S_i$, $\operatorname{Var}\left[\hat{\mu}_{i}\right] \leq \operatorname{Var}\left[\hat{\mu}_{i}^\mathrm{B}\right]$ and thus 
	\begin{equation}\label{halfvar}
		\mu^* + \sqrt{\frac{N-1}{N} \sum_{i}^{N} \operatorname{Var}\left[\hat{\mu}_{i}\right]} \leq  \mu^* + \sqrt{\frac{N-1}{N} \sum_{i}^{N} \operatorname{Var}\left[\hat{\mu}_{i}^\mathrm{B}\right]}.
	\end{equation}
	Therefore, such larger upper bound may cause the maximum value $\mathbb{E}\left[\hat{\mu}_{a_{1}^*}^\mathrm{B}\right]$ to exceed the $\mathbb{E}\left[\hat{\mu}_{\mathrm{SE}}^{*}\right]$.
	Meanwhile, based on the monotonicity in Property 1, it further implies that when number $K$ is too small, the upper of $\mathbb{E}\left[\hat{\mu}_{a_{K}^*}^\mathrm{B}\right]$ tends to be larger than the one of $\mathbb{E}\left[\hat{\mu}_{\mathrm{SE}}^{*}\right]$, which may cause larger overestimation bias.   
	Therefore, the clipping operation guarantees that no matter how small the number of the selected candidates is, the overestimation bias of our estimator is no more than that of the single estimator.
	
	\begin{theorem}  
		Follow the definitions in Theorem 1. 
		If $\operatorname{Var}\left[\min\left\{\hat{\mu}_{a_{(j)}}^{\mathrm{B}},\hat{\mu}_{\mathrm{SE}}^*\right\}\right] \leq \operatorname{Var}\left[\min\left\{\hat{\mu}_{a_{(j+1)}}^{\mathrm{B}},\hat{\mu}_{\mathrm{SE}}^*\right\}\right]$, $1 \leq j < N$, then as the number of action candidates $K$ decreases, $\operatorname{Var}\left[\min\left\{\hat{\mu}_{a_{{K}}^*}^{\mathrm{B}},\hat{\mu}_{\mathrm{SE}}^*\right\}\right]$ decays monotonically, that is $\operatorname{Var}\left[\min\left\{\hat{\mu}_{a_{{K}}^*}^{\mathrm{B}},\hat{\mu}_{\mathrm{SE}}^*\right\}\right] \leq \operatorname{Var}\left[\min\left\{\hat{\mu}_{a_{{K+1}}^*}^{\mathrm{B}},\hat{\mu}_{\mathrm{SE}}^*\right\}\right]$, $1\leq K < N$.
		Instead, if $\operatorname{Var}\left[\min\left\{\hat{\mu}_{a_{(j)}}^{\mathrm{B}},\hat{\mu}_{\mathrm{SE}}^*\right\}\right] \geq \operatorname{Var}\left[\min\left\{\hat{\mu}_{a_{(j+1)}}^{\mathrm{B}},\hat{\mu}_{\mathrm{SE}}^*\right\}\right]$, $1 \leq j < N$, then as the number of action candidates $K$ decreases, the $\operatorname{Var}\left[\min\left\{\hat{\mu}_{a_{{K}}^*}^{\mathrm{B}},\hat{\mu}_{\mathrm{SE}}^*\right\}\right]$ increases monotonically, that is $\operatorname{Var}\left[\min\left\{\hat{\mu}_{a_{{K}}^*}^{\mathrm{B}},\hat{\mu}_{\mathrm{SE}}^*\right\}\right] \geq \operatorname{Var}\left[\min\left\{\hat{\mu}_{a_{{K+1}}^*}^{\mathrm{B}},\hat{\mu}_{\mathrm{SE}}^*\right\}\right]$, $1\leq K < N$.
		%Assume $Var\{\hat{\mu}_{a_{(j)}}^\mathrm{B}\} \leq Var\{\hat{\mu}_{a_{(j+1)}}^\mathrm{B}\}$, $1 \leq j < N$. Then, as the prior preference size $K$ decreases, the $Var\{\hat{\mu}_{a_{K}^*}^\mathrm{B}\}$ decays monotonically, that is $Var\{\hat{\mu}_{a_{K}^*}^\mathrm{B}\} \leq Var\{\hat{\mu}_{a_{K+1}^*}^\mathrm{B}\}$, $1\leq K < N$.
	\end{theorem}
	The proof of Theorem 2 is provided in Appendix I.
	
\subsection{Discussion on  Number of  Candidates}
\label{sec:3c}
As the discussion above, determining a proper number of candidates plays an important role for the low-biased estimation. 
Particularly, when we extend the estimator into the RL field, the online training of the agent requires large amounts of reasonable selections on candidate number $K$  for reliable target value generation. 
To this end, in the following, we analyze and discuss whether we can design an effective selection mechanism that can adaptively determine a good number of candidates. 
In the beginning, we focus on a special case and provide a lemma as follows to demonstrate the relationship between the estimation bias and the candidate number in this case.

\begin{lemma}
If variables $\mathcal{X}=\left\{X_1,\ldots,X_N\right\}$ are i.i.d., the lower bound of the estimation bias of the action candidate based clipped double  estimator satisfies
\begin{equation}
\mathcal{L}\left(\hat{\mu}_{\mathrm{AC}}^*\right)=P\left(\hat{\mu}_{a_{K}^*}^\mathrm{B}>\hat{\mu}^*_{\mathrm{SE}}\right)\mathbb{E}\left[\hat{\mu}^*_{\mathrm{SE}}-\mu^*\right]\geq0.
\end{equation} 
\end{lemma}The proof can be seen in Appendix~I. Based on Lemma 1, for the variables $\mathcal{X}$ without distribution difference, choosing a larger $K$ is reasonable since it has low probability $P\left(\hat{\mu}_{a_{K}^*}^\mathrm{B}>\hat{\mu}^*_{\mathrm{SE}}\right)$ (based on Property 1) which can alleviate the incentive of the positive bias in $\hat{\mu}_{\mathrm{AC}}^*$.  
%and the resulting small overestimated lower bound can largely alleviate the incentive of the positive bias in $\hat{\mu}_{\mathrm{AC}}^*$.  
Similarly, if the distribution difference among the variable set $\mathcal{X}$ is small,  a large number $K$ still be desired for balancing the potential overestimation bias (Appendix~I). 
Extremely, in the case that the variable distributions have significant differences, a smaller number $K$ may be a good choice (Appendix~I).
%Furthermore, when the distributions of the variables in $\mathcal{X}$ have extremely large difference, the single estimator tends to be  unbiased, that is $\mathbb{E}\left[\hat{\mu}^*_{\mathrm{SE}}-\mu^*\right]\approx0$ \cite{hasselt2010double}. Analogously, the lower bound $\mathcal{L}\left(\hat{\mu}_{\mathrm{AC}}^*\right)\approx P\left(\hat{\mu}_{a_{K}^*}^\mathrm{B}\leq \hat{\mu}^*_{\mathrm{SE}} \right)\mathbb{E}\left[\hat{\mu}^*_{\mathrm{DE}}-\mu^*\right]\leq0$. 
%In this case, a smaller size $K$ may be a good choice, since compared to the high $K$, it can get smaller probability $P\left(\hat{\mu}_{a_{K}^*}^\mathrm{B}\leq \hat{\mu}^*_{\mathrm{SE}} \right)$ and the corresponding higher underestimated lower bound $\mathcal{L}\left(\hat{\mu}_{\mathrm{AC}}^*\right)$, which potentially relieves its  underestimation bias. 
Based on the analysis above, we can exploit the distribution difference as the criterion for selecting the proper candidate number.
The larger (smaller) the distribution difference is, the smaller (larger) the number of the candidates is preferred.
%we aim to exploit the distribution difference as the criterion for the selection of $K$.
Specifically, we measure the theoretical distribution difference $\mathcal{H}\left(\mathcal{X}\right)$ with the largest KL-divergence between each pair of the variables, that is $\mathcal{H}\left(\mathcal{X}\right)=\max_{i,j}\operatorname{KL}(\mathcal{I}_i,\mathcal{I}_j)$ where $\mathcal{I}_i$ denotes the unknown distribution of the variable $X_i$. Then, we design a normalization function $\mathcal{J}\left(\mathcal{X}, c\right)$ for mapping the distribution difference $\mathcal{H}\left(\mathcal{X}\right)$ into the range $(0, 1)$:
\begin{equation}
	\begin{split}
		\mathcal{J}\left(\mathcal{X}, c\right)=\frac{1}{1 + \mathcal{H}\left(\mathcal{X}\right)/c},
	\end{split}
\end{equation}
where the hyper-parameter $c>0$ control the sensitivity to the distribution difference $\mathcal{H}\left(\mathcal{X}\right)$. 
%denotes the baseline distribution difference of $\mathcal{H}\left(\mathcal{X}\right)$ and is used to control the normalized value of $\mathcal{H}\left(\mathcal{X}\right)$. 
%A reasonable baseline $c$ requires a similar magnitude as $\mathcal{H}\left(\mathcal{X}\right)$, and too large or too small baseline $c$ promotes the normalized difference to be 1 or 0. 
Finally, we formulate a mapping function $\mathcal{K}\left(\mathcal{X}, c\right)$ as below to bridge the normalized value of $\mathcal{H}\left(\mathcal{X}\right)$ and the number of the candidates:
%The large or small 
%
%If baseline $c$ is  large ($c\gg \mathcal{H}\left(\mathcal{X}\right)$), the normalized value will always  approximate $1$ and similarly the too small $c$ ($c\ll \mathcal{H}\left(\mathcal{X}\right)$) also will lead the normalized value to approximate $0$. Therefore, a reasonable $c$ requires the similar magnitude as $\mathcal{H}\left(\mathcal{X}\right)$. Finally, we build a mapping function $\mathcal{K}\left(\mathcal{X}, c\right)$ as following to bridge the normalized value of $\mathcal{H}\left(\mathcal{X}\right)$ and the number of the candidates:
\begin{equation}\label{kselection11}
	\begin{split}
		\mathcal{K}\left(\mathcal{X}, c\right) = \left\{i\  \Big| \  \frac{i-1}{N}\leq\mathcal{J}\left(\mathcal{X},c\right)<\frac{i}{N}, 1\leq i\leq N\right\},
	\end{split}
\end{equation}
%	\begin{equation}
%	\begin{split}
%		\mathcal{K}\left(\mathcal{X}, c\right) = \sum_{i=1}^{N} i\cdot 1\left\{\frac{i-1}{N}\leq\mathcal{J}\left(\mathcal{X},c\right)<\frac{i}{N}\right\},
%	\end{split}
%\end{equation}
where we equally divide the range $[0,1]$ into $N$ sub-ranges $\big[\frac{i-1}{N}, \frac{i}{N}\big), 1\leq i\leq N$ and each sub-range corresponds to an unique candidate number. 
If $\mathcal{J}\left(\mathcal{X}, c\right)$ falls into the $i$-th sub-range, we utilize $i$ as the candidate number. 
%We utilize the candidate number corresponding to the sub-range that $\mathcal{J}\left(\mathcal{X}, c\right)$ falls into as the final number decision. 
From Eq. \ref{kselection11}, one can see that the larger (smaller) distribution difference tends to choose the smaller (larger) candidate number, which satisfies the analysis above. However, it's impossible to directly calculate the theoretical  $\mathcal{H}\left(\mathcal{X}\right)$ since the distribution of each variable is unknown.
Inspired by \cite{zhang2017weighted},  we assume that the random variables with the largest expected value difference tend to have the largest distribution difference and thus use the maximum sample mean difference to approximate $\mathcal{H}\left(\mathcal{X}\right)$:
\begin{equation}
	\mathcal{H}\left(\mathcal{X}\right) \approx \hat{\mathcal{H}}\left(\mathcal{X}\right)= \hat{\mu}_{a^*}-\hat{\mu}_{a_L},
\end{equation}
where $a^*$ and $a_L$ denote the indexs with the maximum and minimum sample means in $\hat{{\mu}}$, that is $\hat{{\mu}}_{a_L}=\min_i \hat{{\mu}}_i$ and $\hat{{\mu}}_{a^*}=\max_i \hat{{\mu}}_i$. 
Finally, the approximate selection function $ \hat{K}\left(\mathcal{X},c\right)$ is formulated as:
\begin{equation}\label{kselection12}
	\begin{split}
		\hat{\mathcal{K}}\left(\mathcal{X}, c\right) = \left\{i\  \Big| \  \frac{i-1}{N}\leq\hat{\mathcal{J}}\left(\mathcal{X},c\right)<\frac{i}{N}, 1\leq i\leq N\right\},
	\end{split}
\end{equation}
where $\hat{\mathcal{J}}\left(\mathcal{X},c\right)={1}/{(1 + \hat{\mathcal{H}}(\mathcal{X})/{c})}$. Unfortunately, we need manually  determine a proper sensitivity parameter $c$ to fit different task domains since a  fixed $c$ will be difficult to have the same magnitude as $\hat{\mathcal{H}}\left(\mathcal{X},c\right)$ in various domains, and the too large or too small $c$ both may make the normalized value lose the flexibility (approximate 1 or 0).
Despite it, it also should be noted that our selection function in Eq. \ref{kselection12} has the potential to use the same $c$ for the tasks with different settings in the same domain since the magnitudes of $\hat{\mathcal{H}}\left(\mathcal{X},c\right)$  in the same domains tend to be similar. For example, in Section \ref{sec:5}, the same value $c$ can have the best results in almost all settings in the same multi-armed bandit experiment.
%Without the proposed selection function, to achieve the same good results, we have to find different sizes $K$ for different settings. 
Finally, we highlight that in the RL field, a proper $c$ in Eq. \ref{kselection12} can be adaptively determined without the pre-defining and this auto-selection mechanism can bring significant performance gain as shown in the experiment section.
	\section{Action Candidate Based Clipped Double Estimator for Double Q-learning and TD3}
	\label{sec:4}
	In this section, we apply our proposed action candidate based clipped double estimator into Double Q-learning and TD3. 
	For the discrete action task, we first propose the action candidate based clipped Double Q-learning in the tabular setting, and then generalize it to the deep model by using the deep neural network as the Q-value approximator, that is action candidate based clipped Double DQN.
	For the continuous action task, we further combine our estimator with TD3 and form the action candidate based TD3 algorithm.
	%propose a feasible method to approximate the action candidate $\boldsymbol{\mathcal{M}}_K$ and  combine our estimator with TD3, named action candidate based TD3.
	
	\subsection{Action Candidate Based Clipped Double Q-learning}
	\label{sec:4a}
	\textbf{Tabular Version}.
	In tabular setting, action candidate based clipped Double Q-learning stores Q-functions $Q^\mathrm{A}$ and $Q^\mathrm{B}$, and learns them from two separate subsets of the online collected experience. 
	Specifically, after receiving an experience, it randomly chooses one Q-function to update its action value.
	Specifically, in order to update $Q^\mathrm{A}$, we first determine the action candidates:
	% \begin{small}
	\begin{equation} \label{policy}
		\begin{split}
			\mathcal{{M}}_K = \left\{i \mid Q^\mathrm{B}\left(\mathbf{s}', \mathbf{a}_i\right) \in \text{top\ } K \text{\ values in \ }Q^\mathrm{B}\left(\mathbf{s}',\cdot\right) \right\}.
		\end{split}
	\end{equation}According to the action value function $Q^\mathrm{A}$, the action $\mathbf{a}_K^*$ is the maximal valued action among $\mathcal{{M}}_K$ at state $\mathbf{s}'$.
	Then, we update $Q^\mathrm{A}$ via the target value as below:
	\begin{equation} \label{target_acq}
		\begin{split}
			y^{\mathrm{AC\_CDQ}}=r + \gamma \min\left\{Q^\mathrm{B}\left(\mathbf{s}', \mathbf{a}_K^*\right), {\max}_{\mathbf{a}} Q^\mathrm{A}\left(\mathbf{s}', \mathbf{a}\right)\right\}.
		\end{split}
	\end{equation}
	During the training process, the explored action is calculated with $\epsilon$-greedy exploration strategy based on the action values $Q^\mathrm{A}$ and $Q^\mathrm{B}$.  Note that in the tabular version, the number of action candidates balances the overestimation in Q-learning and the underestimation in clipped Double Q-learning. 
	
Furthermore, in order to adapatively select a proper candidate number $K$ for each target construction in Eq. \ref{target_acq} during training, similar to the selection function $\hat{K}\left(\mathcal{X}, c\right)$ used in our estimator, we propose a corresponding selection function $\hat{K}'\left(Q^\mathrm{A},\mathbf{s}',\bar{c}\right)$ for our tabular variant as follows:
\begin{equation}\label{select_acq}
	\begin{split}
		\hat{K}'\left(Q^\mathrm{A},\mathbf{s}',\bar{c}\right) = \left\{i\  \Big| \  \frac{i-1}{N}\leq\hat{\mathcal{J}}\left(Q^\mathrm{A},\mathbf{s}'\right)<\frac{i}{N}, 1\leq i\leq N\right\},
	\end{split}
\end{equation}
%	\begin{equation} \label{select_acq}
%	\begin{split}
%		\hat{K}'\left(Q^\mathrm{A},\mathbf{s}',\bar{c}\right) = \sum_{i=1}^{N} i\cdot 1\left\{\frac{i-1}{N}\leq\frac{\bar{c}}{\bar{c} + \hat{\mathcal{H}}\left(Q^\mathrm{A},\mathbf{s}'\right)}<\frac{i}{N}\right\},
%	\end{split}
%\end{equation}
where $\hat{\mathcal{J}}\left(Q^\mathrm{A},\mathbf{s}'\right)=1 / (1 + \hat{\mathcal{H}}\left(Q^\mathrm{A},\mathbf{s}'\right) / \bar{c})$ and $\hat{\mathcal{H}}\left(Q^\mathrm{A},\mathbf{s}'\right)= Q^\mathrm{A}\left(\mathbf{s}',\mathbf{a}^*\right)-Q^\mathrm{A}\left(\mathbf{s}',\mathbf{a}_L\right)$. $\mathbf{a}^*$ and $\mathbf{a}_L$ corresponds to the indexs with the maximum and minimum action values in $Q^\mathrm{A}\left(\mathbf{s}', \cdot\right)$, respectively. Notably, in the RL setting, we use the historical average value as the current reference value denoted as  $\bar{c}$. If the estimated distribution difference $\hat{\mathcal{H}}\left(Q^\mathrm{A},\mathbf{s}'\right)$ is larger than the average value, we think the distribution difference is large and tend to choose the small size $K$ and vice versa. It is analogous to the idea of the advantage function in the value function based RL field. Specifically, we store $\hat{\mathcal{H}}\left(Q^\mathrm{A},\mathbf{s}'\right)$ in each time-step and utilize the mean of the lastest $M$ values as $\bar{c}$ (we set $M$ to 50 in our experiments). More details are shown in Algorithm~\ref{alg:algorithm1}.

Finally, we prove  that the action candidate based clipped Double Q-learning can converge to the optimal policy in the finite MDP setting. The details can be seen in Appendix II. 

\begin{theorem}
	The action candidate based clipped Double Q-learning can converge to the optimal policy when the following conditions are satisfied:\\
	1) Each state action pair is sampled an infinite number of times. \\
	2) The MDP is finite, that is $|\mathcal{S}\times \mathcal{A}|< \infty$.\\
	3) $\gamma \in [0, 1)$.\\
	4) Q values are stored in a lookup table.\\
	5) Both $Q^\mathrm{A}$ and $Q^\mathrm{B}$ receive an infinite number of updates.\\
	6) The learning rates satisfy $\alpha_{t}\left(\mathbf{s}, \mathbf{a}\right) \in[0,1], \sum_{t} \alpha_{t}\left(\mathbf{s}, \mathbf{a}\right)=\infty, \sum_{t}\left(\alpha_{t}\left(\mathbf{s}, \mathbf{a}\right)\right)^{2}<\infty$ with probability 1 and $\alpha_t\left(\mathbf{s}, \mathbf{a}\right)=0, \forall\left(\mathbf{s}, \mathbf{a}\right) \neq\left(\mathbf{s}_{t}, \mathbf{a}_{t}\right)$.\\
	7) $\operatorname{Var}[r\left(\mathbf{s}, \mathbf{a}\right)]<\infty, \forall \mathbf{s}, \mathbf{a}$.
\end{theorem}

\textbf{Deep Version.}
For the task with the high-dimensional sensory input, we further propose the deep version of action candidate based clipped Double Q-learning, named action candidate based clipped Double DQN. In our framework, we maintain two deep Q-networks $Q\left(\mathbf{s},\mathbf{a};\boldsymbol{\theta}_1\right)$ and $Q\left(\mathbf{s},\mathbf{a};\boldsymbol{\theta}_2\right)$, and an experience buffer $\mathcal{D}$. In each time-step, we update one Q-function, such as $Q\left(\mathbf{s},\mathbf{a};\boldsymbol{\theta}_1\right)$, with the loss function as below:
\begin{equation}
	\mathcal{L} =\mathbb{E}_{\mathcal{D}}\left[\left(y^{\mathrm{AC\_CDDQN}}-Q\left(\mathbf{s},\mathbf{a};\boldsymbol{\theta}_1\right)\right)^2\right].
\end{equation}
The target value $y^{\mathrm{AC\_CDDQN}}$ is defined as:
\begin{equation}
	y^{\mathrm{AC\_CDDQN}}=r + \gamma \min\left\{Q\left(\mathbf{s},\mathbf{a}^*_K;\boldsymbol{\theta}_2\right),\max_{\mathbf{a}}Q\left(\mathbf{s},\mathbf{a};\boldsymbol{\theta}_1\right)\right\},
\end{equation}where $a^*_K = \arg\max_{\mathbf{a}\in\mathcal{M}_K}Q\left(\mathbf{s},\mathbf{a};\boldsymbol{\theta}_1\right)$ and $\mathcal{M}_K$ is the action candidates with the top $K$ values in $Q\left(\mathbf{s},\cdot;\boldsymbol{\theta}_2\right)$. Similar to the tabular version, the number of action candidates can  balance the overestimation in DQN and the underestimation in clipped Double DQN.

		\begin{figure*}[ht]
		\centering
		\includegraphics[width=\textwidth]{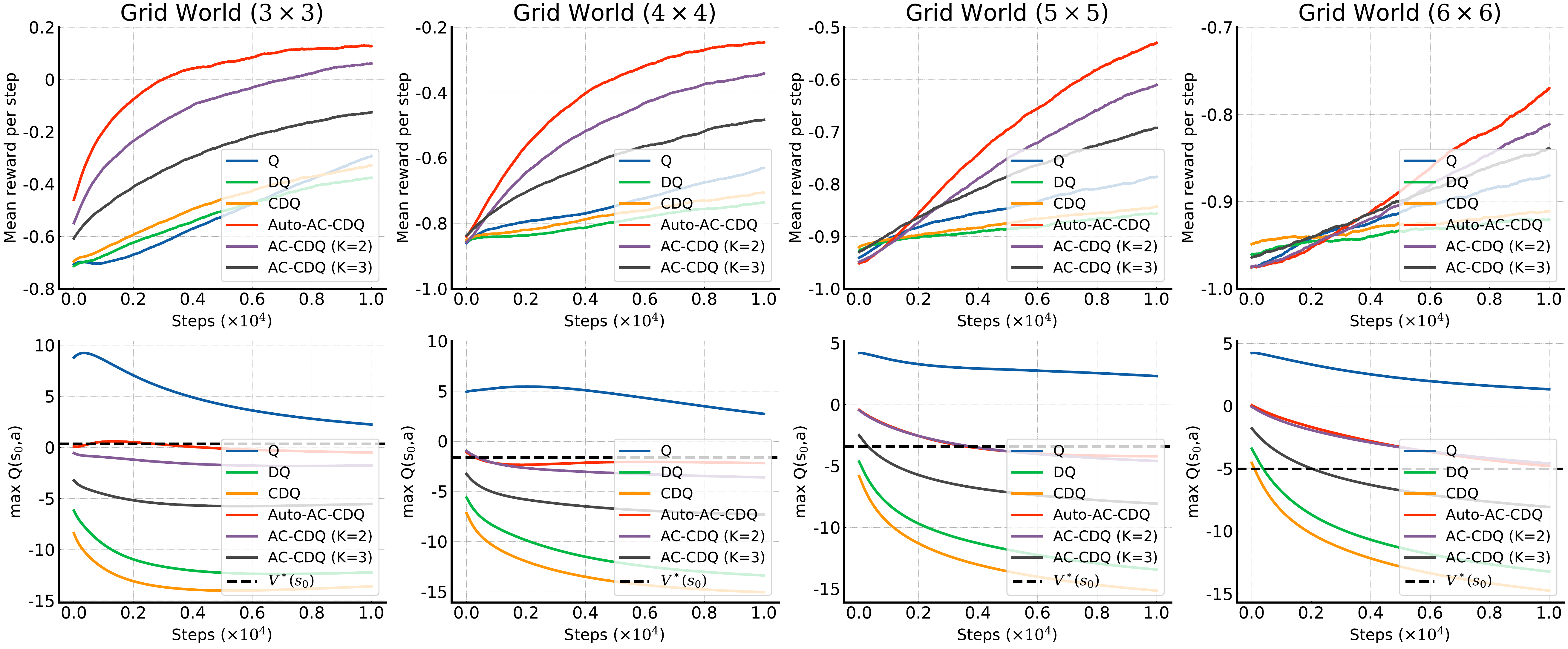}
		\caption{The top plots show the mean reward per step and the bottom plots show the estimated maximum action value from the state $s_0$ (the black dash line demotes the optimal state value $V^*(s_0)$). The results are averaged over 10000 experiments and each experiment contains 10000 steps. 
			We set the number of action candidates to 2 and 3, respectively. Q: Q-learning, DQ: Double Q-learning, CDQ: clipped Double Q-learning.
		}	
		\label{fig:gridworld1}
	\end{figure*}
	\subsection{Action Candidate Based TD3}
	\label{sec:4b}
	As shown in Algorithm 2, the algorithm framework for the continuous action task follows the design in TD3.
	To approximate the optimal action values, we construct two Q-networks $Q\left(\mathbf{s}, \mathbf{a};\boldsymbol{\theta}_1\right)$ and $Q\left(\mathbf{s}, \mathbf{a};\boldsymbol{\theta}_2\right)$ and two target Q-networks $Q\left(\mathbf{s}, \mathbf{a};\boldsymbol{\theta}_1^-\right)$ and $Q\left(\mathbf{s}, \mathbf{a};\boldsymbol{\theta}_2^-\right)$.
	In addition, two deterministic policy networks $\mu\left(\mathbf{s}; {\boldsymbol{\phi}_1}\right)$ and $\mu\left(\mathbf{s}; {\boldsymbol{\phi}_2}\right)$, and two target networks $\mu\left(\mathbf{s}; {\boldsymbol{\phi}_1^-}\right)$ and $\mu\left(\mathbf{s}; {\boldsymbol{\phi}_2^-}\right)$ are exploited to represent the optimal decisions corresponding to  $Q\left(\mathbf{s}, \mathbf{a};\boldsymbol{\theta}_1\right)$, $Q\left(\mathbf{s}, \mathbf{a};\boldsymbol{\theta}_2\right)$,  $Q\left(\mathbf{s}, \mathbf{a};\boldsymbol{\theta}_1^-\right)$ and $Q\left(\mathbf{s}, \mathbf{a};\boldsymbol{\theta}_2^-\right)$.
	
	Due to the continuity of the actions, it is impossible to precisely determine the top $K$ action candidates $\mathcal{M}_K$ like in the discrete action case. We first exploit our deterministic policy network $\mu\left(\mathbf{s}'; {\boldsymbol{\phi}_2^-}\right)$ to approximate the global optimal action $\mathbf{a}^* = \arg\max_{\mathbf{a}} Q\left(\mathbf{s}', \mathbf{a};\boldsymbol{\theta}_2^-\right)$. Based on the estimated global optimal action $\mathbf{a}^*$, we randomly select $K$ actions ${\mathcal{M}}_{K}$ in the $\delta$-neighborhood of $\mathbf{a}^*$ as the action candidates. 
	Specifically, we draw $K$ samples from a Gaussian distribution $\mathcal{N}\left(\mu\left(\mathbf{s}'; {\boldsymbol{\phi}_2^-}\right), \bar{\boldsymbol{\sigma}}\right)$:
	% \begin{small}
	\begin{equation} \label{policy}
		\begin{split}
			{\mathcal{M}}_K = \left\{ \mathbf{a}_i \mid \mathbf{a}_i \sim \mathcal{N}\left(\mu\left(\mathbf{s}'; {\boldsymbol{\phi}_2^-}\right), \bar{\boldsymbol{\sigma}}\right), i = 1, \ldots ,K  \right\},
		\end{split}
	\end{equation}
	% \end{small}
	where the hyper-parameter $\bar{\boldsymbol{\sigma}}$ is the standard deviation. Both Q-networks  are updated via the following target value:
	% \begin{small}
	\begin{equation} \label{policy}
		\begin{split}
			&y^{\mathrm{AC\_TD3}}= \\
			&\ \ r + \gamma \min\big\{Q\left(\mathbf{s}', \mathbf{a}_K^*;\boldsymbol{\theta}_2^-\right),Q\left(\mathbf{s}', \mu\left(\mathbf{s}'; {\boldsymbol{\phi}_1^-}\right);\boldsymbol{\theta}_1^-\right)\big\},
			%y = r + \gamma \min\left\{Q_{\theta_2'}(s', a_K^*), Q_{\theta_1'}(s', \pi_{\theta_1'}(s'))\right\},
		\end{split}
	\end{equation}
	% \end{small}
	where $\mathbf{a}_K^* = \arg\max_{\mathbf{a} \in {\mathcal{M}}_K} Q\left(\mathbf{s}', \mathbf{a};\boldsymbol{\theta}_1^-\right)$. %and the target value is also clipped by $Q\left(s', \mu\left(s'; {\boldsymbol{\phi}_1^-}\right);\boldsymbol{\theta}_1^-\right)$.
	The parameters of two policy networks are updated along the direction that can improve their corresponding Q-networks.
	For more details please refer to Algorithm~\ref{actd3}.
	
	It should be noted that, in each time-step, action candidate based clipped Double Q-learning just randomly updates only one Q-function (random updating).
	Instead, action candidate based TD3 exploits the same target value to update both Q-functions at the same time. In the following, we provide the theorem that such simultaneous updating in the tabular case still retrains convergence as  Theorem 3.
	\begin{theorem}
		For simultaneous updating in the tabular case, that is both Q-functions are updated with the same target value $y_t$ in each time step:
		\begin{equation}
			\begin{split}
				&Q_{t+1}^{\mathrm{A}}\left(\mathbf{s}_{t}, \mathbf{a}_{t}\right) \leftarrow Q_{t}^{\mathrm{A}}\left(\mathbf{s}_{t}, \mathbf{a}_{t}\right)+\alpha_{t}\left(\mathbf{s}_{t}, \mathbf{a}_{t}\right)\left(y_{t}-Q_{t}^{\mathrm{A}}\left(\mathbf{s}_{t}, \mathbf{a}_{t}\right)\right) \\
				&Q_{t+1}^{\mathrm{B}}\left(\mathbf{s}_{t}, \mathbf{a}_{t}\right) \leftarrow Q_{t}^{\mathrm{B}}\left(\mathbf{s}_{t}, \mathbf{a}_{t}\right)+\alpha_{t}\left(\mathbf{s}_{t}, \mathbf{a}_{t}\right)\left(y_{t}-Q_{t}^{\mathrm{B}}\left(\mathbf{s}_{t}, \mathbf{a}_{t}\right)\right),
			\end{split}
		\end{equation}
		where $y_{t}=r_{t}+\gamma \min \left\{Q^{\mathrm{B}}\left(\mathbf{s}_{t+1}, \mathbf{a}_{K}^{*}\right), Q^{\mathrm{A}}\left(\mathbf{s}_{t+1}, \mathbf{a}^{*}\right)\right\}$, it can converge to the optimal policy when the same conditions in Theorem 3  are satisfied.
	\end{theorem}
	The proof is provided in Appendix II.
	
		\begin{figure*}[ht]
		\centering
		\includegraphics[width=\textwidth]{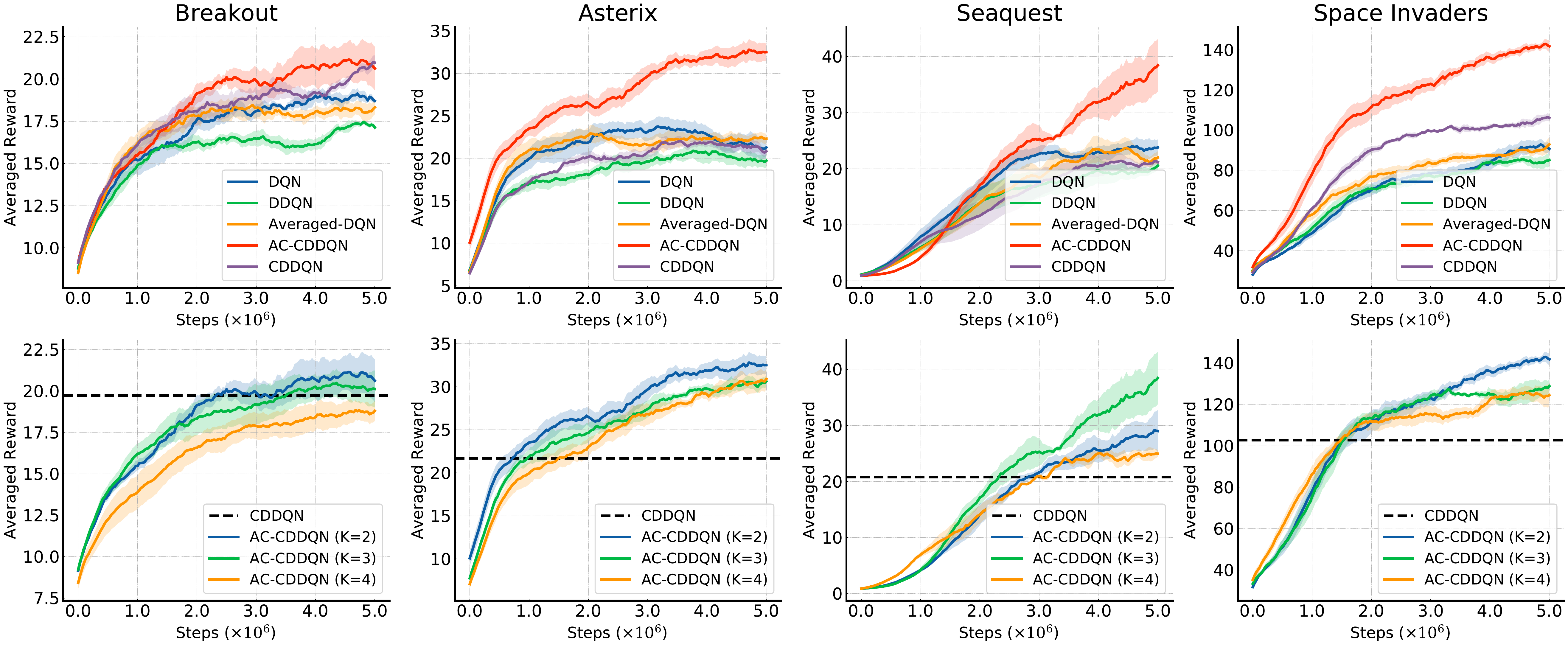}
		\caption{Learning curves on the four MinAtar benchmark environments. The results are averaged over five independent learning trials and the shaded area represents half a standard deviation. DDQN: Double DQN, CDDQN: clipped Double DQN.
		}	
		\label{fig:minatari}
	\end{figure*}
	
	\begin{figure}[]
		\centering  
		\subfigure[Number of impressions ($\times10^4$)]{
			\label{Fig.sub.1}
			\includegraphics[width=\columnwidth]{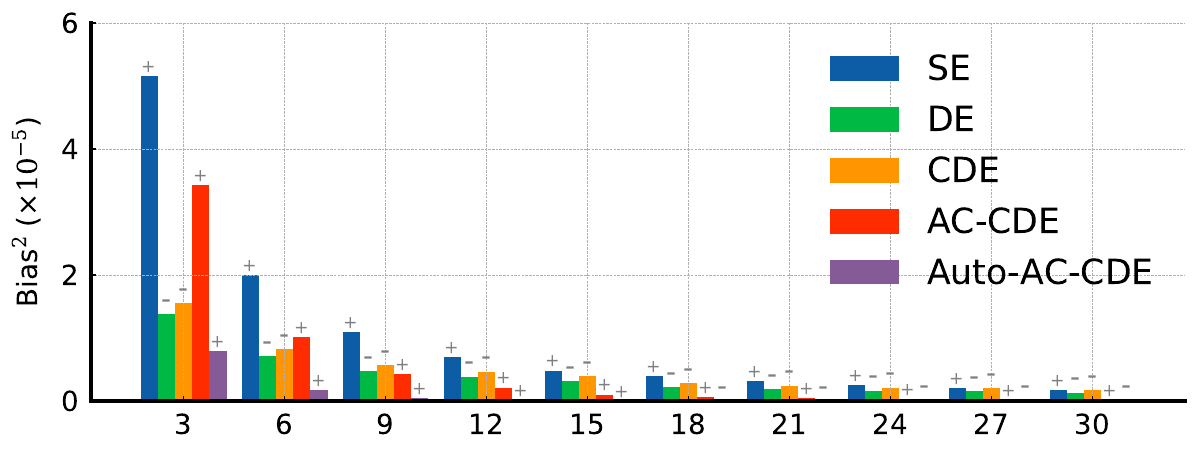}}
		\subfigure[Number of ads]{
			\label{Fig.sub.2}
			\includegraphics[width=\columnwidth]{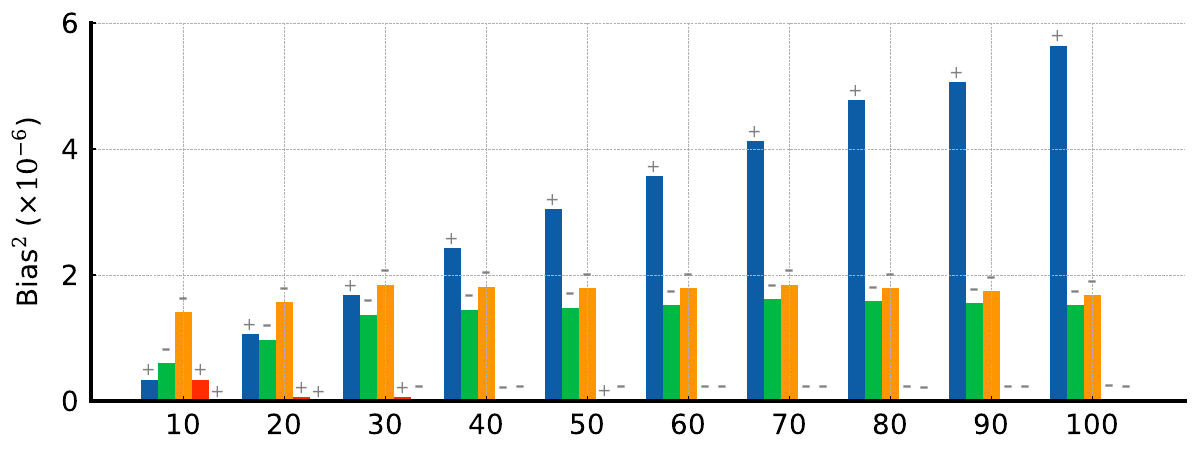}}
		\subfigure[Max probability in range ($\times 10^{-2}$)]{
			\label{Fig.sub.3}
			\includegraphics[width=\columnwidth]{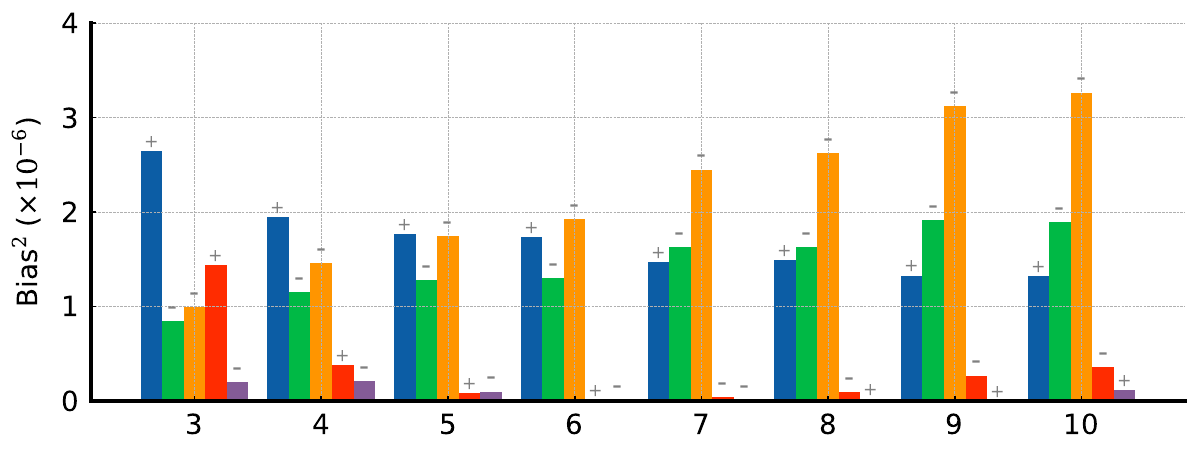}}
		\caption{Comparison on the multi-armed bandits for internet ads in three cases: (a) Varying the number of impressions; (b) Varying the number of ads; (c) Varying the max probability. 
			The symbol on the bar represents the sign of the bias.
			The results are averaged over $2,000$ experiments. 
			We use $15\%$ of actions as the action candidates.		
			The hyper-parameter $c$ of selection function in Auto-AC-CDE is set to $0.005$.
			SE: single estimator, DE: double estimator, CDE: clipped double estimator. %(Adaptive-)AC-CDE: (adaptive) action candidate based clipped double estimator.
		}
		\label{fig:games}
	\end{figure}
			\begin{figure*}[t]
		\centering
		\includegraphics[width=1\textwidth]{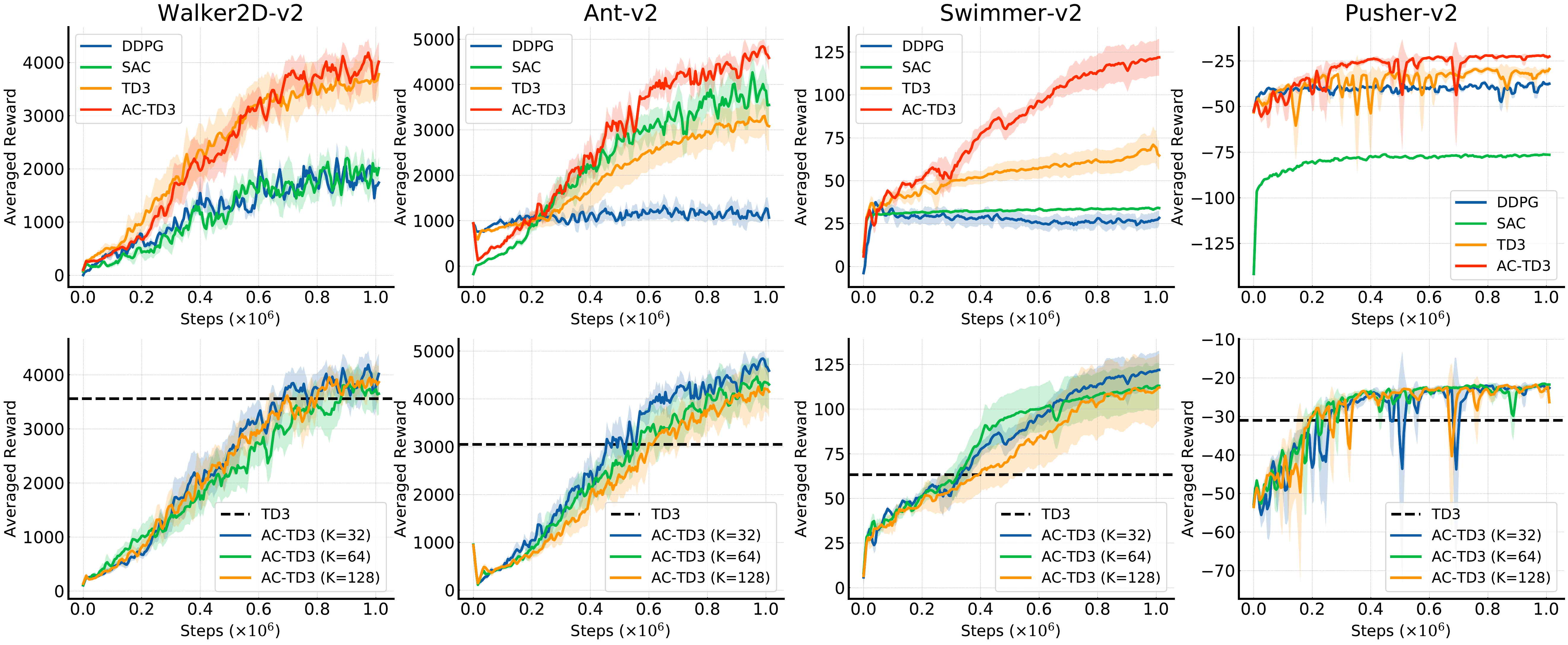}
		\caption{Top row: Learning curves for the OpenAI Gym continuous control tasks. The shaded region represents half a standard deviation of the average evaluation over 10 trials. Bottom row: Learning curves for AC-TD3 with different numbers of the action candidates ($K=\left\{32, 64, 128\right\}$).
			%Bottom plots: the estimation of the expected return with respect to the initial state $s_0$ of the game. The dash lines represent the real discounted return.
		}	
		\label{fig:mujoco}
	\end{figure*}
	\section{Experiments}
	\label{sec:5}
	In this section, we  empirically evaluate our method on the discrete and continuous action tasks. For the discrete action tasks, we conduct the following three experiments: 
	\begin{itemize}
		\item For action candidate based clipped double estimator (AC-CDE) and its adaptive version with the selection function (Auto-AC-CDE), we compare them with single estimator \cite{hasselt2010double}, double estimator \cite{hasselt2010double} and clipped double estimator \cite{fujimoto2018addressing} on the multi-armed bandits problem.
		\item For action candidate based clipped Double Q-learning (AC-CDQ) and its adaptive variant with the selection function (Auto-AC-CDQ), we compare them with Q-learning \cite{watkins1992q}, Double Q-learning \cite{hasselt2010double} and clipped Double Q-learning \cite{fujimoto2018addressing} on the grid world environment.
		\item For action candidate based clipped Double DQN (AC-CDDQN), we compare them with DQN \cite{mnih2015human}, Double DQN \cite{van2016deep}, Averaged-DQN \cite{anschel2017averaged} and clipped Double DQN \cite{fujimoto2018addressing}  on several benchmark games in MinAtar \cite{young2019minatar}.
	\end{itemize}
	
	For the continuous action tasks, we compare our action candidate based TD3 (AC-TD3) with TD3 \cite{fujimoto2018addressing}, SAC \cite{haarnoja2018soft} and DDPG \cite{lillicrap2015continuous} on six MuJoCo \cite{todorov2012mujoco} based benchmark tasks implemented in OpenAI Gym \cite{dhariwal2017openai}.
	
%	\subsection{Implementation Detials}
%For action candidate based clipped Double DQN, the number of frames is $5\cdot10^6$; the discount factor is $0.99$; reward scaling is 1.0; the batch size is $32$; the buffer size is $1\cdot10^6$; the frequency of updating the target network is $1000$; the optimizer is RMSprop with learning $2.5\cdot10^{-4}$, squared gradient momentum $0.95$ and minimum squared gradient $0.01$; the iteration per time step is $1$. For action candidate Based TD3, the number of frames is $1\cdot10^6$; the discount factor is $0.99$; reward scaling is 1.0; the batch size is $256$; the buffer size is $1\cdot10^6$; the frequency of updating the target network is $2$; the optimizers for actor and critic are Adams with learning $3\cdot10^{-4}$; the iteration per time step is $1$. All experiments are conducted on a server with NVIDIA TITAN V. 
		\begin{figure}[]
		\centering
		\includegraphics[width=1\columnwidth]{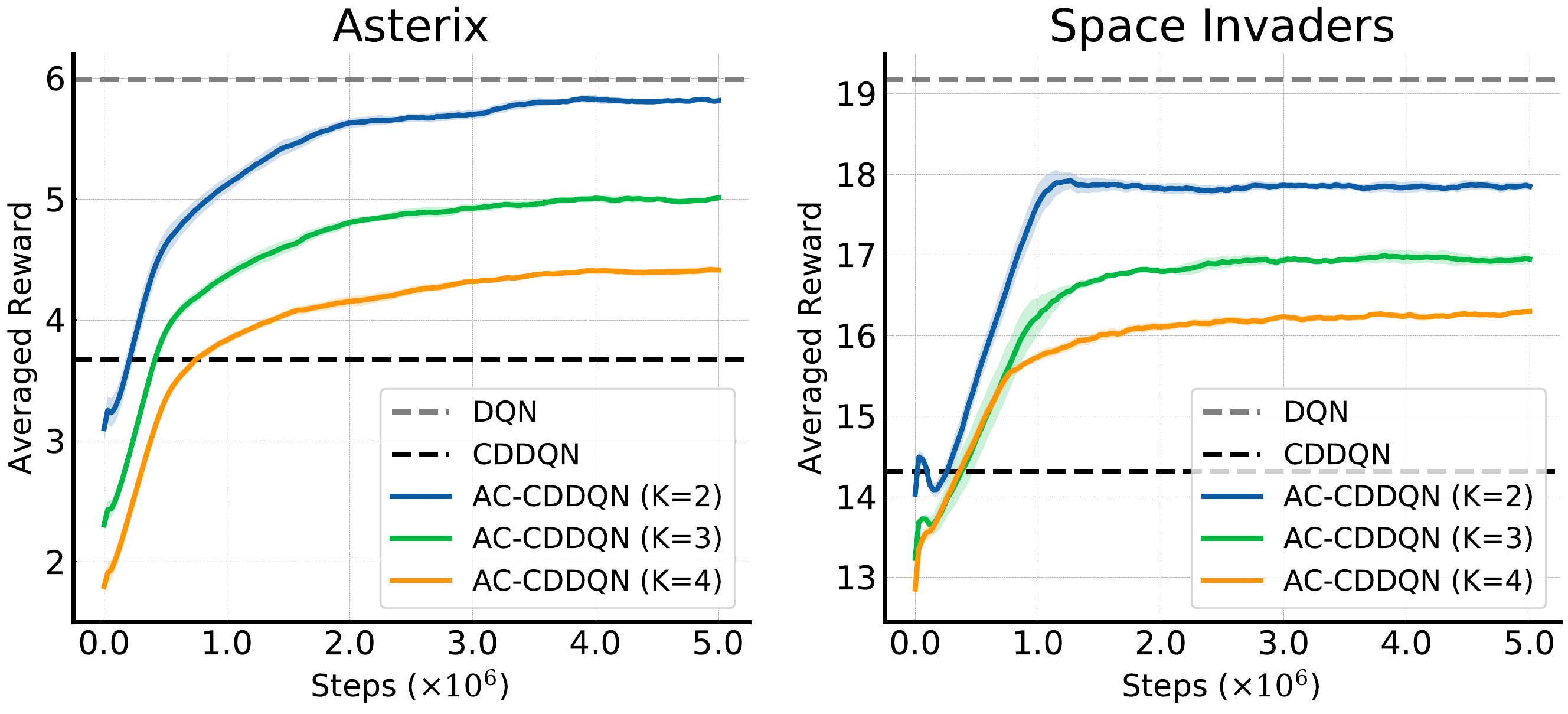}
		\caption{Learning curves about the estimated maximum action values for AC-CDDQN with different numbers of the action candidates (K=2, 3, 4). CDDQN: clipped Double DQN.
			%CDDQN: clipped Double DQN.
		}	
		\label{fig:minatari_vary_k}
	\end{figure}
	\subsection{Discrete Action Tasks}
	\label{sec:5a}
	\textbf{Multi-Armed Bandits For Internet Ads.} 
	In this experiment, we employ the framework of the multi-armed bandits to choose the best ad to show on the website among a set of $M$ possible ads, each one with an unknown fixed expected return per visitor.
	For simplicity, we assume each ad has the same return per click, such that the best ad is the one with the maximum click rate. We model the click event per visitor in each ad $i$ as the Bernoulli event with mean $m_i$ and variance $(1-m_i)m_i$. In addition, all ads are assumed to have the same visitors, which means that given $N$ visitors, $N/M$ Bernoulli experiments will be executed to estimate the click rate of each ad. The default configuration in our experiment is $N=30,000$, $M=30$ and the mean click rates uniformly sampled from the interval $\left[0.02, 0.05\right]$. Based on this configuration, there are three settings: (1) We vary the number of visitors $N=\left\{30,000, 60,000,\ldots,270,000, 300,000\right\}$. (2) We vary the number of ads $M=\left\{10, 20,\ldots,90,100\right\}$. (3) We vary the upper limit of the sampling interval of mean click rate (the original is $0.05$) with values $\left\{0.03, 0.04,\ldots,0.09,0.1\right\}$.
	%modify the interval of the mean click rates by changing this interval's 
	%upper limit (the original is 0.5) 
	%value of the upper limit with values in $\left\{0.03, 0.04,\ldots,0.09,0.1\right\}$.
	
	To compare the absolute bias, we evaluate the single estimator, double estimator, clipped double estimator, AC-CDE, and Auto-AC-CDE with the square of bias ($bias^2$) in each setting.
As shown in Fig.~\ref{fig:games}, compared to other estimators, AC-CDE and Auto-AC-CDE own the lowest $bias^2$ in almost all experimental settings. 
It mainly benefits from the effective balance of our proposed estimator between the overestimation bias of the single estimator and the underestimation bias of the clipped double estimator.
Moreover, AC-CDE has the lower $bias^2$ than single estimator in all cases while in some cases it has the larger $bias^2$ than clipped double estimator such as the leftmost columns in Fig.~\ref{fig:games} (a) and (c). It's mainly due to that although AC-CDE can reduce the underestimation bias of clipped double estimator, too small number of action candidates may also in turn cause overestimation bias. % (no more than the single estimator). 
Thus, the absolute value of such overestimation bias may be larger than the one of the underestimation bias in the clipped double estimator. Despite this, AC-CDE can guarantee that the positive bias is no more than the single estimator and the negative bias is also no more than the clipped double estimator. 
In addition, owing to the proposed selection function for action size $K$, Auto-AC-CDE obtains the best results on all settings. Particularly, in the leftmost two columns in the top plot in Fig. \ref{fig:games}, the absolute values of the estimation biases in AC-CDE are larger than the double estimator and the clipped double estimator. Instead, the proper selection on size $K$ in Auto-AC-CDE effectively avoids it and brings the lowest biases.

	\textbf{Grid World.} 
	As a MDP task, in a $N\times N$ grid world, there are total $N^2$ states. The starting state $s_0$ is in the lower-left cell and the goal state is in the upper-right cell. Each state has four actions: east, west, south and north.  At any state, moving to an adjacent cell is deterministic, but a collision with the edge of the world will result in no movement. Taking an action at any state will receive a random reward which is set as below: if the next state is not the goal state, the random reward is $-6$ or $+4$ and if the agent arrives at the goal state, the random reward is $-30$ or $+40$. 
	For the reward criterion, since the minimum number of the actions for reaching the goal step is $2\times N - 1$, the optimal average reward per action is $\frac{5-2(N-1)}{2\times N -1}=\frac{7-2\times N}{2\times N -1}$.
	For the bias criterion, with the discount factor $\gamma$, the optimal value of the maximum value action in the starting state $s_0$ is $5\gamma^{2(N-1)}-\sum_{i=0}^{2N-3}\gamma^i$. 
	We set $N=\left\{3,4,5,6\right\}$ to construct our grid world environments and compare the Q-learning, Double Q-learning, clipped Double Q-learning, AC-CDQ ($K=2,3$) and Auto-AC-CDQ on the mean reward per step and estimation error (see Fig.~\ref{fig:gridworld1}). 
	
	From the top plots, one can see that AC-CDQ ($K=2,3$) and Auto-AC-CDQ can obtain the higher mean reward than other methods in all tested environments.
	We further plot the estimation error about the optimal state value $V^*\left(\mathbf{s}_0\right)$ in bottom plots. 
	Compared to Q-learning, Double Q-learning, and clipped Double Q-learning, AC-CDQ ($K=2,3$) and Auto-AC-CDQ also show the much lower estimation bias (more closer to the dash line), which means that it can better assess the action value and thus help generate more valid action decision.
	Moreover, our AC-CDQ can significantly reduce the underestimation bias in clipped Double Q-learning.
	Notably, as demonstrated in Theorem 1, 
	the underestimation bias in the case of $K=2$ is smaller than that in the case of $K=3$.
	%the smaller number of the action candidates ($K=2$) can cause larger underestimation reduction than the case of $K=3$, 
	And AC-CDQ can effectively balance the overestimation bias in Q-learning and the underestimation bias in clipped Double Q-learning. 
	Despite it, the fixed candidate size in AC-CDQ isn't able to adaptively generate proper target value for each experience, which potentially degrades its performance and action value estimation. Instead, benefitting from the adaptive candidate size selection in Auto-AC-CDQ, compared to AC-CDQ, it can obtain significant reward gain and bias reduction. Particularly, in the bottom plot of Fig. \ref{fig:gridworld1}, Auto-AC-CDQ achieves the almost unbiased estimation in all environments.
	%
	%One can see that the large estimatiom bias 
	%
	%
	%It mainly owes to that AC-DQ ($K=2,3$) can obtain the more precise estimation about the optimal state value.
	%As shown in bottom plots, the estimation curves of AC-CDQ ($K=2,3$) are more close to the $V^*(s_0)$ than Q, DQ and CDQ.
	%
	%It mainly owes to that AC-DQ ($K=2,3$) can estimate the optimal state value more precisely, which can be reflected by the bottom plots where the learning curves of AC-CDQ ($K=2,3$) are more close to the $V^*(s_0)$ than Q and DQ. Moreover, the especially accurate estimation of AC-DQ ($K=2$) brings it the superior performance advantage than other methods.
	%Notably, as demonstrated in Theorem 2, the smaller number of the action candidates ($K=2$) can cause larger underestimation reduction than the case of $K=3$, and AC-DQ can effectively balance the overestimation bias in Q and the underestimation bias in DQ.
%	\begin{figure}[h]
%		\centering
%		\includegraphics[width=1\columnwidth]{./pics/plot_gridworld_experiment.pdf}
%		\caption{
%			The top plots show the mean reward per step and the bottom plots show the estimated maximum action value from the state $s_0$ (the black dash line demotes the optimal state value $V^*(s_0)$). The results are averaged over 10000 experiments and each experiment contains 10000 steps. 
%			We set the number of the action candidates to 2 and 3, respectively. 
%			%Q: Q-learning, DQ: Double Q-learning, CDQ: clipped Double Q-learning.
%		}	
%		\label{fig:gridworld1}
%	\end{figure}

	\textbf{MinAtar.}
	MinAtar is a game platform for testing the reinforcement learning algorithms, which uses a simplified state representation to model the game dynamics of Atari from ALE \cite{bellemare2013arcade}.
	In this experiment, we compare the performance of DQN, Double DQN, Averaged-DQN, clipped Double DQN and AC-CDDQN on four released MinAtar games including Breakout, Asterix, Seaquest, and Space Invaders.
	We exploit the convolutional neural network as the function approximator and use the game image as the input to train the agent in an end-to-end manner. 
	Following the settings in \cite{young2019minatar}, we set the batch size and the model updating frequency to $32$ and $1$, respectively. 
	 The discounted factor, initial exploration, and the final exploration are set to $0.99$, $0.00025$, and $0.1$, respectively. For the experience buffer, the replay memory size and the replay start size are set to $100,000$ and $5,000$.
	We use the RMSProp as the optimizer with the learning rate $0.00025$, gradient momentum $0.95$, and the minimum squared gradient $0.01$ for model training. All experimental results are obtained after 5M frames.
%	the hyper-parameters and settings of our method are set as below: the batch size is 32; the replay memory size is 100,000; the update frequency is 1; the discounted factor is 0.99; the learning rate is 0.00025; the initial exploration is 1; the final exploration is 0.1; the replay start size is 5,000. We use the RMSProp as the optimizer with the gradient momentum 0.95 and the minimum squared gradient 0.01 for model training. The experimental results are obtained after 5M frames.
	
	%The optimizer is RMSProp with the gradient momentum 0.95 and the minimum squared gradient 0.01. The experimental results are obtained after 5M frames.
	
	The top plots in Fig.~\ref{fig:minatari} represent the training curve about the averaged reward of each algorithm. 
	It shows that compared to DQN, Double DQN, Averaged-DQN, and clipped Double DQN, AC-CDDQN can obtain better or comparable performance while they have similar convergence speeds in all four games. 
	Especially, for Asterix, Seaquest, and Space Invaders, AC-CDDQN can achieve significantly higher averaged rewards compared to the clipped Double DQN and obtain the gains of $36.3\%$, $74.4\%$ and $19.8\%$, respectively. 
	The great performance gain mainly owes to that AC-CDDQN can effectively balance the overestimation bias in DQN and the underestimation bias in clipped Double DQN.
	Moreover, as shown in the bottom plots in  Fig.~\ref{fig:minatari}, we also test the averaged rewards of different numbers of action candidates $K=\left\{2,3,4\right\}$ for AC-CDDQN. The plots show that compared to the clipped Double DQN (the dash line), the action candidate mechanism in AC-CDDQN can almost consistently bring the robust and superior performance with different action candidate sizes (except for the case of candidate size $4$ in the Breakout game), which further verifies the effectiveness of our proposed improvement. Moreover, as shown in the plots of Fig. \ref{fig:minatari_vary_k}, as the size of the action candidates decreases, its estimated maximum action value keeps increasing, which means the corresponding underestimation bias reduces monotonically. It means that our deep variant also empirically follows the monotonicity property in Theorem 1.
%	 our deep version can effectively balance the overestimated DQN and the underestimated clipped Double DQN. Further, it also
%	empirically follows the monotonicity in Theorem 1, that is as the number $K$ of action candidates decreases, the underestimation bias in clipped Double DQN reduces monotonically.
%		\begin{figure}[t]
%	\centering
%	\includegraphics[width=1\columnwidth]{./pics/plot_score_TNNSL.pdf}
%	\caption{Top row: Learning curves for the OpenAI Gym continuous control tasks. The shaded region represents half a standard deviation of the average evaluation over 10 trials. 
%		%Bottom plots: the estimation of the expected return with respect to the initial state $s_0$ of the game. The dash lines represent the real discounted return.
%	}	
%	\label{fig:mujoco}
%\end{figure}
	\begin{table}[]
	%\caption{After the 200 episodes the average performance with the largest scores for  DQN, DDQN and AC-DDQN.}	
	\centering
	\renewcommand{\arraystretch}{1.3}
	\resizebox{1\linewidth}{!}{
		\begin{tabular}{c|ccccc}
			\toprule[1.2pt]
			& AC-TD3  & TD3 & SAC & DDPG \\
			\hline\hline
			Pusher   & \textbf{-22.7 $\pm$ 0.39}  & -31.8 & -76.7 & -38.4\\
			Reacher & \textbf{-3.5 $\pm$ 0.06}  & \textbf{-3.6} & -12.9 & -8.9 \\
			Walker2d &  \textbf{3800.3 $\pm$ 130.95}  & 3530.4 & 1863.8 & 1849.9 \\
			Hopper  & 2827.2 $\pm$ 83.2 & {2974.8} & \textbf{3111.1} & 2611.4 \\
			Swimmer &  \textbf{116.2 $\pm$ 3.63} & 63.2 & 33.6 & 30.2 \\
			Ant &  \textbf{4391 $\pm$ 205.6} & 3044.6 & 3646.5 & 1198.64 \\\bottomrule[1.2pt]
	\end{tabular}}
	\vspace{1pt}
	\caption{Averaged rewards over last 30\% episodes during training process.}
	\label{mujoco_table}
\end{table}

\begin{figure}[]
	\centering
	\includegraphics[width=1\columnwidth]{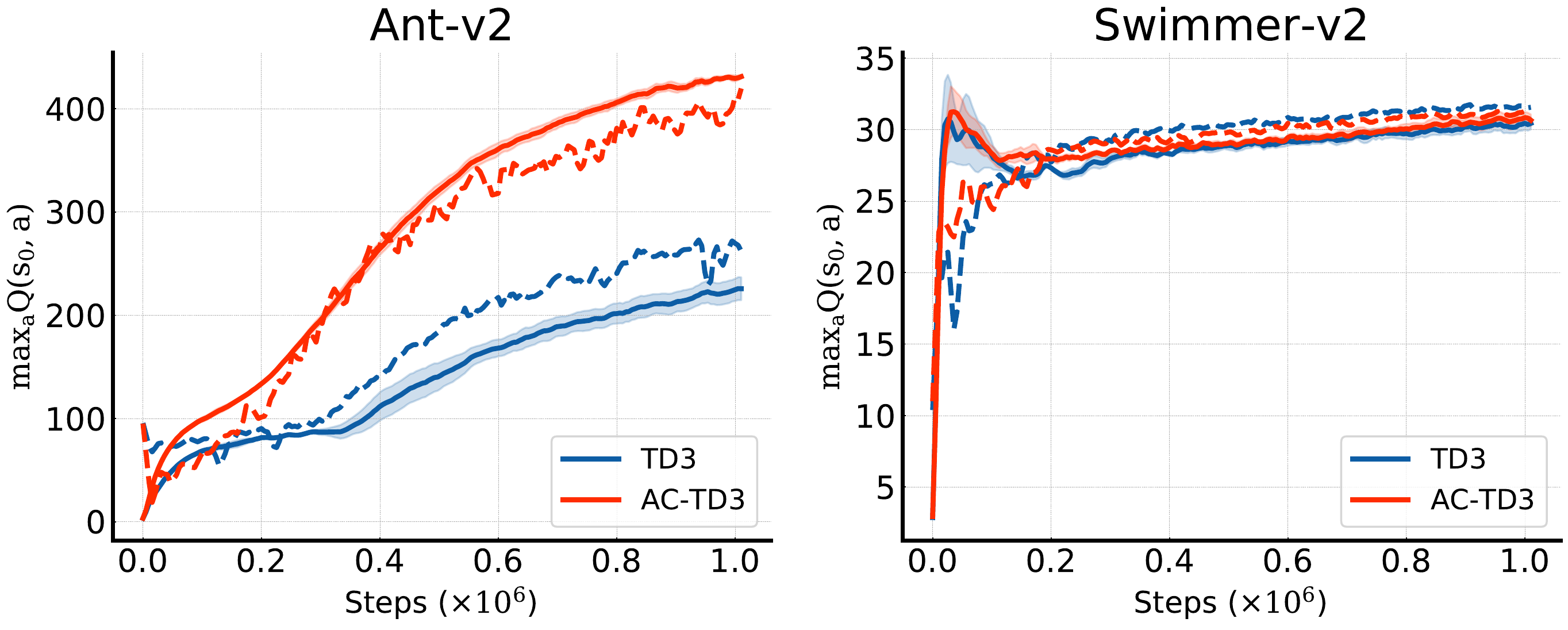}
	\caption{The estimation of the expected return with respect to the initial state $\mathbf{s}_0$ of the game. The dash lines represent the real discounted return.
	}	
	\label{fig:mujoco_bias}
\end{figure}
	
	\subsection{Continuous Action Task}
	\textbf{MuJoCo Tasks.}
	We verify our variant for continuous action, AC-TD3, on six MuJoCo continuous control tasks from OpenAI Gym including Ant-v2, Walker2D-v2, Swimmer-v2, Pusher-v2, Hopper-v2 and Reacher-v2.
	%The screenshots of the environments can be seen in Fig.~\ref{fig:games}.
	We compare our method against the DDPG and two state of the art methods: TD3 and SAC.
	In our method, we exploit the TD3 as our baseline and just modify it with our action candidate mechanism.	The implementation details of our method is set as below: the number of the action candidate is set to $32$;  
	the number of frames is $1\cdot10^6$; the discount factor is $0.99$; reward scaling is 1.0; the batch size is $256$; the buffer size is $1\cdot10^6$; the frequency of updating the target network is $2$; the optimizers for actor and critic are Adams with learning $3\cdot10^{-4}$; the iteration per time step is $1$. %All experiments are conducted on a server with NVIDIA TITAN V. 
	We run all tasks with 1 million timesteps and the trained policies are evaluated every $5,000$ timesteps. 
	
	We list the training curves of Walker2D-v2, Ant-v2, Swimmer-v2, and Pusher-v2 in the top row of Fig.~\ref{fig:mujoco} and the comprehensive comparison results are listed in Table~\ref{mujoco_table}.
	From Table~\ref{mujoco_table}, one can see that DDPG performs poorly in most environments, and TD3 and SAC can't handle some tasks such as Swimmer-v2 well.
	In contrast, AC-TD3 consistently obtains robust and competitive performance in all environments.
	Particularly, AC-TD3 owns comparable learning speeds across all tasks and can achieve higher averaged reward than TD3 (our baseline) in most environments except for Hopper-v2. 
	Such significant performance gain verifies that our proposed approximate action candidate method in the continuous action case is effective empirically. 
	Furthermore, we test the robustness of our method on different sizes of the action candidates $K=\left\{32, 64, 128\right\}$ and the training curves are presented in the bottom plots of Fig.~\ref{fig:mujoco}.  The results demonstrate that the  AC-TD3  can obtain superior scores than TD3 for each action candidate size.
	Moreover, we also explain the performance advantage of our AC-TD3 over TD3 from the perspective of the bias in Fig.~\ref{fig:mujoco_bias}.
	The plots show that in Ant-v2 and Swimmer-v2, AC-TD3 tends to have a lower estimation bias than TD3 about the expected return with regard to the initial state $\mathbf{s}_0$, which potentially helps the agent assess the action at some state better and then generate the more reasonable policy.
%	\begin{figure}[t]
%		\centering
%		\includegraphics[width=1\columnwidth]{./pics/plot_score_bias.pdf}
%		\caption{Top row: Learning curves for the OpenAI Gym continuous control tasks. The shaded region represents half a standard deviation of the average evaluation over 10 trials. Bottom plots: the estimation of the expected return with respect to the initial state $s_0$ of the game. The dash lines represent the real discounted return.
%		}	
%		\label{fig:mujoco}
%	\end{figure}
	
	\section{Conclusion}
	\label{section:6}
	In this paper, we proposed an action candidate based clipped double estimator to approximate the maximum expected value. Furthermore, we applied this estimator to form the action candidate based clipped Double Q-learning. Theoretically, the underestimation bias in clipped Double Q-learning decays monotonically as the number of action candidates decreases. The number of action candidates can also control the trade-off between overestimation and underestimation. 
Finally, we also extend our clipped Double Q-learning to the deep version and the continuous action tasks.
	Experimental results demonstrate that our proposed methods yield competitive performance.

	\bibliographystyle{IEEEtran}
	\bibliography{IEEEfull}

\begin{thebibliography}{10}
\providecommand{\url}[1]{#1}
\csname url@samestyle\endcsname
\providecommand{\newblock}{\relax}
\providecommand{\bibinfo}[2]{#2}
\providecommand{\BIBentrySTDinterwordspacing}{\spaceskip=0pt\relax}
\providecommand{\BIBentryALTinterwordstretchfactor}{4}
\providecommand{\BIBentryALTinterwordspacing}{\spaceskip=\fontdimen2\font plus
\BIBentryALTinterwordstretchfactor\fontdimen3\font minus
  \fontdimen4\font\relax}
\providecommand{\BIBforeignlanguage}[2]{{%
\expandafter\ifx\csname l@#1\endcsname\relax
\typeout{** WARNING: IEEEtran.bst: No hyphenation pattern has been}%
\typeout{** loaded for the language `#1'. Using the pattern for}%
\typeout{** the default language instead.}%
\else
\language=\csname l@#1\endcsname
\fi
#2}}
\providecommand{\BIBdecl}{\relax}
\BIBdecl

\bibitem{thrun2000reinforcement}
S.~Thrun and M.~L. Littman, ``Reinforcement learning: {A}n introduction,''
  \emph{AI Magazine (2000)}, vol.~21, no.~1, pp. 103--103.

\bibitem{kaelbling1996reinforcement}
L.~P. Kaelbling, M.~L. Littman, and A.~W. Moore, ``Reinforcement learning: {A}
  survey,'' \emph{Journal of artificial intelligence research (1996)}, vol.~4,
  pp. 237--285.

\bibitem{watkins1989learning}
C.~J. C.~H. Watkins, ``Learning from delayed rewards,'' 1989.

\bibitem{arjona2018rudder}
J.~A. Arjona-Medina, M.~Gillhofer, M.~Widrich, T.~Unterthiner, J.~Brandstetter,
  and S.~Hochreiter, ``Rudder: {R}eturn decomposition for delayed rewards,''
  \emph{arXiv preprint arXiv:1806.07857 (2018)}.

\bibitem{szepesvari2010algorithms}
C.~Szepesv{\'a}ri, ``Algorithms for reinforcement learning,'' \emph{Synthesis
  lectures on artificial intelligence and machine learning (2010)}, vol.~4,
  no.~1, pp. 1--103.

\bibitem{wu2021deep}
Y.~Wu, S.~Liao, X.~Liu, Z.~Li, and R.~Lu, ``Deep reinforcement learning on
  autonomous driving policy with auxiliary critic network,'' \emph{IEEE
  transactions on neural networks and learning systems}, 2021.

\bibitem{watkins1992q}
C.~J. Watkins and P.~Dayan, ``Q-learning,'' \emph{Machine learning (1992)},
  vol.~8, no. 3-4, pp. 279--292.

\bibitem{bertsekas1995neuro}
D.~P. Bertsekas and J.~N. Tsitsiklis, ``Neuro-dynamic programming: an
  overview,'' in \emph{Proceedings of 1995 34th IEEE conference on decision and
  control (1995)}.

\bibitem{thrun1993issues}
S.~Thrun and A.~Schwartz, ``Issues in using function approximation for
  reinforcement learning,'' in \emph{Proceedings of the 1993 Connectionist
  Models Summer School (1993)}.

\bibitem{szita2008many}
I.~Szita and A.~L{\H{o}}rincz, ``The many faces of optimism: {A} unifying
  approach,'' in \emph{ICML (2008)}.

\bibitem{strehl2009reinforcement}
A.~L. Strehl, L.~Li, and M.~L. Littman, ``Reinforcement learning in finite
  {MDP}s: {PAC} analysis,'' \emph{Journal of Machine Learning Research (2009)},
  vol.~10, no. Nov, pp. 2413--2444.

\bibitem{van2018deep}
H.~Van~Hasselt, Y.~Doron, F.~Strub, M.~Hessel, N.~Sonnerat, and J.~Modayil,
  ``Deep reinforcement learning and the deadly triad,'' \emph{arXiv preprint
  arXiv:1812.02648 (2018)}.

\bibitem{van2011insights}
H.~P. van Hasselt, \emph{Insights in reinforcement learning}.\hskip 1em plus
  0.5em minus 0.4em\relax Hado van Hasselt (2011).

\bibitem{xue2021inverse}
W.~Xue, B.~Lian, J.~Fan, P.~Kolaric, T.~Chai, and F.~L. Lewis, ``Inverse
  reinforcement q-learning through expert imitation for discrete-time
  systems,'' \emph{IEEE Transactions on Neural Networks and Learning Systems},
  2021.

\bibitem{strehl2006pac}
A.~L. Strehl, L.~Li, E.~Wiewiora, J.~Langford, and M.~L. Littman, ``Pac
  model-free reinforcement learning,'' in \emph{ICML (2006)}.

\bibitem{riedmiller2005neural}
M.~Riedmiller, ``Neural fitted {Q} iteration--first experiences with a data
  efficient neural reinforcement learning method,'' in \emph{ECML (2005)}, pp.
  317--328.

\bibitem{tosatto2017boosted}
S.~Tosatto, M.~Pirotta, C.~d’Eramo, and M.~Restelli, ``Boosted fitted
  {Q}-iteration,'' in \emph{ICML (2017)}, pp. 3434--3443.

\bibitem{ernst2005tree}
D.~Ernst, P.~Geurts, and L.~Wehenkel, ``Tree-based batch mode reinforcement
  learning,'' \emph{Journal of Machine Learning Research (2005)}, vol.~6, pp.
  503--556.

\bibitem{abed2018double}
B.~H. Abed-alguni and M.~A. Ottom, ``Double delayed {Q}-learning,''
  \emph{International Journal of Artificial Intelligence (2018)}, vol.~16,
  no.~2, pp. 41--59.

\bibitem{mnih2013playing}
V.~Mnih, K.~Kavukcuoglu, D.~Silver, A.~Graves, I.~Antonoglou, D.~Wierstra, and
  M.~Riedmiller, ``Playing atari with deep reinforcement learning,''
  \emph{arXiv preprint arXiv:1312.5602 (2013)}.

\bibitem{mnih2015human}
V.~Mnih, K.~Kavukcuoglu, D.~Silver, A.~A. Rusu, J.~Veness, M.~G. Bellemare,
  A.~Graves, M.~Riedmiller, A.~K. Fidjeland, G.~Ostrovski \emph{et~al.},
  ``Human-level control through deep reinforcement learning,'' \emph{nature
  (2015)}, vol. 518, no. 7540, pp. 529--533.

\bibitem{sorokin2015deep}
I.~Sorokin, A.~Seleznev, M.~Pavlov, A.~Fedorov, and A.~Ignateva, ``Deep
  attention recurrent {Q}-network,'' \emph{arXiv preprint arXiv:1512.01693
  (2015)}.

\bibitem{fan2020theoretical}
J.~Fan, Z.~Wang, Y.~Xie, and Z.~Yang, ``A theoretical analysis of deep
  {Q}-learning,'' in \emph{Learning for Dynamics and Control (2020)}, pp.
  486--489.

\bibitem{lee2013bias}
D.~Lee, B.~Defourny, and W.~B. Powell, ``Bias-corrected {Q}-learning to control
  max-operator bias in {Q}-learning,'' in \emph{2013 IEEE Symposium on Adaptive
  Dynamic Programming and Reinforcement Learning (2013)}.

\bibitem{lee2019bias}
D.~Lee and W.~B. Powell, ``Bias-corrected {Q}-learning with multistate
  extension,'' \emph{IEEE Transactions on Automatic Control (2019)}, vol.~64,
  no.~10, pp. 4011--4023.

\bibitem{song2019revisiting}
Z.~Song, R.~Parr, and L.~Carin, ``Revisiting the softmax bellman operator:
  {N}ew benefits and new perspective,'' in \emph{ICML (2019)}.

\bibitem{d2016estimating}
C.~D'Eramo, M.~Restelli, and A.~Nuara, ``Estimating maximum expected value
  through gaussian approximation,'' in \emph{ICML (2016)}.

\bibitem{cini2020deep}
A.~Cini, C.~D'Eramo, J.~Peters, and C.~Alippi, ``Deep reinforcement learning
  with weighted {Q}-learning,'' \emph{arXiv preprint arXiv:2003.09280 (2020)}.

\bibitem{anschel2017averaged}
O.~Anschel, N.~Baram, and N.~Shimkin, ``Averaged-{DQN}: {V}ariance reduction
  and stabilization for deep reinforcement learning,'' in \emph{ICML (2017)}.

\bibitem{lan2020maxmin}
Q.~Lan, Y.~Pan, A.~Fyshe, and M.~White, ``Q-learning,'' \emph{arXiv preprint
  arXiv:2002.06487 (2020)}.

\bibitem{hasselt2010double}
H.~V. Hasselt, ``Double {Q}-learning,'' in \emph{NIPS (2010)}.

\bibitem{van2013estimating}
H.~van Hasselt, ``Estimating the maximum expected value: {A}n analysis of
  (nested) cross validation and the maximum sample average,'' \emph{arXiv
  preprint arXiv:1302.7175 (2013)}.

\bibitem{zhang2017weighted}
Z.~Zhang, Z.~Pan, and M.~J. Kochenderfer, ``Weighted double {Q}-learning.'' in
  \emph{IJCAI (2017)}.

\bibitem{chen2021randomized}
X.~Chen, C.~Wang, Z.~Zhou, and K.~Ross, ``Randomized ensembled double
  {Q}-learning: {L}earning fast without a model,'' \emph{arXiv preprint
  arXiv:2101.05982 (2021)}.

\bibitem{xiong2020finite}
H.~Xiong, L.~Zhao, Y.~Liang, and W.~Zhang, ``Finite-time analysis for double
  {Q}-learning,'' \emph{NIPS (2020)}.

\bibitem{fujimoto2018addressing}
S.~Fujimoto, H.~Van~Hoof, and D.~Meger, ``Addressing function approximation
  error in actor-critic methods,'' \emph{arXiv preprint arXiv:1802.09477
  (2018)}.

\bibitem{van2016deep}
H.~Van~Hasselt, A.~Guez, and D.~Silver, ``Deep reinforcement learning with
  double q-learning,'' in \emph{AAAI (2016)}.

\bibitem{young2019minatar}
K.~Young and T.~Tian, ``Minatar: An atari-inspired testbed for more efficient
  reinforcement learning experiments,'' \emph{arXiv preprint arXiv:1903.03176},
  2019.

\bibitem{haarnoja2018soft}
T.~Haarnoja, A.~Zhou, K.~Hartikainen, G.~Tucker, S.~Ha, J.~Tan, V.~Kumar,
  H.~Zhu, A.~Gupta, P.~Abbeel \emph{et~al.}, ``Soft actor-critic algorithms and
  applications,'' \emph{arXiv preprint arXiv:1812.05905 (2018)}.

\bibitem{lillicrap2015continuous}
T.~P. Lillicrap, J.~J. Hunt, A.~Pritzel, N.~Heess, T.~Erez, Y.~Tassa,
  D.~Silver, and D.~Wierstra, ``Continuous control with deep reinforcement
  learning,'' \emph{arXiv preprint arXiv:1509.02971 (2015)}.

\bibitem{todorov2012mujoco}
E.~Todorov, T.~Erez, and Y.~Tassa, ``Mujoco: {A} physics engine for model-based
  control,'' in \emph{IROS (2012)}.

\bibitem{dhariwal2017openai}
P.~Dhariwal, C.~Hesse, O.~Klimov, A.~Nichol, M.~Plappert, A.~Radford,
  J.~Schulman, S.~Sidor, Y.~Wu, and P.~Zhokhov, ``Openai baselines,'' 2017.

\bibitem{bellemare2013arcade}
M.~G. Bellemare, Y.~Naddaf, J.~Veness, and M.~Bowling, ``The arcade learning
  environment: {A}n evaluation platform for general agents,'' \emph{Journal of
  Artificial Intelligence Research (2013)}, vol.~47, pp. 253--279.

\end{thebibliography}

\end{document}